%% file: bert-pse.tex
\documentclass[11pt,a4paper]{article}
\usepackage[hyperref]{emnlp2020}
\usepackage{times}
\usepackage{latexsym}

\usepackage{url}
\usepackage{microtype}

\usepackage[T1]{fontenc}
\usepackage{xcolor}
\usepackage{graphicx}
\usepackage{caption}
\usepackage{subcaption}
\usepackage{amsfonts}
\usepackage{amssymb}
\usepackage{enumerate}
\usepackage{algorithm}
\usepackage{algorithmicx}
\usepackage[noend]{algpseudocode}
\usepackage{booktabs}
\usepackage{multirow}
\usepackage{amsmath}
\usepackage{xspace}

\aclfinalcopy %
\def\aclpaperid{1040} %

\def\ptl{pretrained language model\xspace}
\def\ptls{pretrained language models\xspace}

\def\Ptls{Pretrained language models\xspace}

\newcounter{notecounter}
\newcommand{\enotesoff}{\long\gdef\enote##1##2{}}
\newcommand{\enoteson}{\long\gdef\enote##1##2{{
\stepcounter{notecounter}
{\large\bf \hspace{1cm}\arabic{notecounter} $<<<$ ##1: ##2 $>>>$\hspace{1cm}}}}}
\enoteson
\enotesoff

\def\figref#1{Figure~\ref{fig:#1}}
\def\figlabel#1{\label{fig:#1}\label{p:#1}}

\def\Tabref#1{Table~\ref{tab:#1}}
\def\tabref#1{Table~\ref{tab:#1}}
\def\tablabel#1{\label{tab:#1}\label{p:#1}}

\def\secref#1{\S\ref{sec:#1}}
\def\seclabel#1{\label{sec:#1}}
\def\eqref#1{Eq.~\ref{eqn:#1}}

\title{Quantifying the Contextualization of Word Representations\\
with Semantic Class Probing}

\author{
  Mengjie Zhao\textsuperscript{†}, 
  Philipp Dufter\textsuperscript{†}, 
  Yadollah Yaghoobzadeh\textsuperscript{‡},
  Hinrich Sch\"{u}tze\textsuperscript{†}\\
  \textsuperscript{†} CIS, LMU Munich, Germany \ 
  \textsuperscript{‡} Microsoft Turing, Montréal, Canada\\
  {\tt mzhao@cis.lmu.de}
}

\date{}

\begin{document}
\maketitle
\begin{abstract}
  \Ptls achieve state-of-the-art results on many NLP
  tasks, but there are still many open
  questions about how and why they work so well.
  We investigate the contextualization of words in BERT.
  We quantify the amount of contextualization, i.e., how
  well words are interpreted in context, by studying
  the extent to which \emph{semantic classes} of a word can be
  inferred from its contextualized embedding.
  Quantifying  contextualization helps in understanding
  and utilizing \ptls.
  We show that
  the top layer
  representations support highly accurate
  inference of semantic classes; that
  the strongest contextualization effects occur in the lower
  layers; that
  local context is mostly sufficient for contextualizing words;
  and
  that top layer representations are more task-specific
  after finetuning while lower layer representations are more
  transferable.  Finetuning uncovers task-related features,
  but pretrained knowledge about contextualization is still well preserved.
\end{abstract}

\section{Introduction}
\Ptls like ELMo \citep{peters-etal-2018-deep},
BERT \citep{devlin-etal-2019-bert}, and
XLNet \citep{Yang2019XLNetGA}
are top performers in NLP
because they learn contextualized representations,
i.e., representations that reflect the interpretation of a word in context as
opposed to its general meaning, which is less helpful in solving
NLP tasks.
As stated,
\ptls contextualize words, is clear \emph{qualitatively}; there has
been little work
on investigating
contextualization, i.e., to which extent a word can be interpreted in context,
\emph{quantitatively}.

We use BERT \citep{devlin-etal-2019-bert}
as our \ptl and quantify contextualization by investigating
how well BERT infers \textbf{semantic classes} (s-classes) of a word
in context,
e.g., the s-class \emph{organization} for ``Apple" in ``Apple stock
rises" vs.\ the s-class \emph{food} in ``Apple juice is
healthy".
We use s-class inference as a proxy for contextualization since
accurate s-class inference reflects a successful
contextualization of a word: an effective interpretation of
the word
in context.

We adopt the methodology of probing
\citep{adi2016fine, shi-etal-2016-string,belinkov-etal-2017-evaluating,
  liu-etal-2019-linguistic, tenney2019you, belinkov-glass-2019-analysis,
hewitt-liang-2019-designing,  yaghoobzadeh-etal-2019-probing}:
diagnostic classifiers are applied to \ptl embeddings
to determine whether they encode desired syntactic or semantic features.

By probing for s-classes we quantify directly
where and how contextualization happens in BERT.
E.g., we find that the strongest
contextual interpretation effects occur in the lower
layers and that the top two layers contribute little to
contextualization. We also investigate how the amount of
context available affects contextualization.

In addition, since \ptls in practice need
to be finetuned on downstream tasks \citep{devlin-etal-2019-bert,
peters-etal-2019-tune}, we further investigate the interactions between
finetuning and contextualization. We show that the pretrained knowledge
about contextualization is well preserved in finetuned models.

We make the following \textbf{contributions}:
(i) We investigate how
accurately BERT interprets words in context.
We find that BERT's performance is high (almost 85\% $F_1$), but
that there is still room for improvement.
(ii) We quantify how much each additional layer in BERT
contributes to contextualization.
We find that the strongest contextual
interpretation effects occur in the lower layers. 
The top two layers
seem to be
optimized only for the
pretraining objective of
predicting masked words \citep{devlin-etal-2019-bert} and
only add small increments to
contextualization.
(iii) We investigate the amount of context BERT needs to
exploit for interpreting a word and find that BERT
effectively integrates local context
up to five words to the left and to the right (a
10-word context window).
(iv) We investigate the dynamics of BERT's representations
in finetuning.
We find that finetuning has little effect on
lower layers, suggesting that they
are more easily transferable across tasks.
Higher layers are strongly changed for word-level tasks
like part-of-speech tagging,
but less noticeably for sentence-level tasks like paraphrase
classification.
Finetuning uncovers task-related features, but the
knowledge captured in pretraining is well preserved.
We quantify these effects by s-class inference performance.

\begin{table}
  \centering\small
  \begin{tabular}{|c|c|}
    \hline
    GloVe     & BERT     \\ \hline
    suits     & suits    \\
    lawsuit   & suited   \\
    filed     & lawsuit  \\
    lawsuits  & \#\#suit \\
    sued      & lawsuits \\
    complaint & slacks   \\
    jacket    & 47th     \\ \hline
  \end{tabular}
  \caption{Nearest neighbors of ``suit'' in GloVe and
    in 
    BERT (BERT-base-uncased)
  wordpiece embeddings}
  \tablabel{suitexample}
\end{table}

\section{Motivation and Methodology}
\seclabel{method}
The key benefit of \ptls \citep{mccann2017learned,
  peters-etal-2018-deep, radford2019language,
  devlin-etal-2019-bert} is that they produce
contextualized embeddings that are useful in NLP.
The top layer contextualized word representations from \ptls are
widely utilized; however, the fact that \ptls implement \emph{a process of}
contextualization -- starting with a completely
uncontextualized layer of wordpieces at the bottom --
is not well studied.
\Tabref{suitexample} gives an example: BERT's wordpiece
embedding of ``suit'' is not contextualized: it contains
several
meanings of the word, including ``to suit'' (``be convenient''),
lawsuit, and garment (``slacks'').
Thus, there is no difference
in this respect between BERT's wordpiece embeddings and
uncontextualized word embeddings
like GloVe \citep{Pennington2014}.
\Ptls start out with an uncontextualized representation
at the lowest layer, then
gradually contextualize it. This is the process
we analyze in this paper.

For investigating the contextualization process, one
possibility is to use word senses
and to tap resources
like the
WordNet
(WN) \citep{wordnetcite}  based word sense disambiguation benchmarks of
the Senseval series \citep{edmonds-cotton-2001-senseval,
  snyder-palmer-2004-english, raganato-etal-2017-word}. However, the abstraction
level in WN sense inventories has been criticized as too fine-grained
\citep{izquierdo2009empirical}, providing limited information to applications
requiring higher level abstraction. Various
levels of granularity of abstraction have been explored such as WN domains
\citep{magnini-cavaglia-2000-integrating}, supersenses
\citep{ciaramita2003supersense, levine2019sensebert} and basic level concepts
\citep{izquierdo2007exploring}.
In this paper, we use semantic classes
(s-classes) \citep{yarowsky-1992-word, resnik-1993-semantic,kohomban-lee-2005-learning,
  yaghoobzadeh-etal-2019-probing} as the proxy for the meaning contents of words
to study the contextualization capability of BERT.
Specifically, we use the Wikipedia-based resource for Probing Semantics
in Word Embeddings (Wiki-PSE)
\citep{yaghoobzadeh-etal-2019-probing} which is detailed in
\secref{psestats}.

\begin{table}[]
  \centering\footnotesize
  \begin{tabular}{r|rrr}

          & words  & comb's & contexts  \\ \hline
    train & 35,399 & 62,184 & 2,178,895 \\
    dev   & 8,850  & 15,437 & 542,938   \\
    test  & 44,250 & 77,706 & 2,722,893
  \end{tabular}
  \caption{Number of words, word-s-class combinations, and contexts
    per split in our probing dataset. Appendix \secref{psedetails}
    shows the 34 s-classes and statistics per class.}
  \tablabel{datastas}
\end{table}

\section{Probing Dataset and Task}
\subsection{Probing dataset}
\seclabel{psestats}
For s-class probing,
we use the s-class labeled corpus Wiki-PSE
\citep{yaghoobzadeh-etal-2019-probing}.
It consists of a set of 34 s-classes, an inventory of
word$\rightarrow$s-class mappings and
an English  Wikipedia text
corpus in which words in context are labeled with the
34 s-classes. For example, contexts of ``Apple''
that refer to the company are labeled with
``organization''.
We refer to a word labeled with an s-class as a
word-s-class combination, e.g.,
``@apple@-organization''.\footnote{In Wiki-PSE, s-class-labeled
  occurrences are enclosed with ``@'', e.g.,
  ``@apple@''.}

The Wiki-PSE text corpus contains $>$550 million tokens,
$>$17 million of which are
annotated with an s-class. Working on the entire
Wiki-PSE with BERT is not feasible, e.g., the
word-s-class combination
``@france@-location'' has 98,582 contexts.
Processing all these contexts by BERT consumes significant amounts of energy
\citep{strubell-etal-2019-energy,schwartz2019green}
and time. Hence for each
word-s-class combination,
we sample a maximum of 100 contexts
to speed up our
experiments.
Wiki-PSE provides a balanced train/test split; we use 20\% of
the training set as our development set.
\Tabref{datastas} gives
statistics of our dataset.

\subsection{Probing for semantic classes}
\seclabel{probingexperiments}
For each of the 34 s-classes in Wiki-PSE, we train a binary classifier to
diagnose if an input embedding encodes information for inferring the
s-class.

\subsubsection{Probing uncontextualized embeddings}
\seclabel{probetypelevelsetup}
We make a distinction in this paper
between
two different factors that contribute to
BERT's  performance:
(i) a powerful learning architecture that gives rise to
high-quality representations
and (ii) contextualization in applications, i.e.,
words are represented as contextualized embeddings for solving NLP
tasks.
Here,     we adopt \citet{schuster-etal-2019-cross}'s method of
computing uncontextualized BERT embeddings
(AVG-BERT-$\ell$, see \secref{representationlearners})
and show that
(i) alone already has a strong positive effect on
performance when compared to other uncontextualized embeddings.
So BERT's representation learning yields high
  performance, even when used in a completely uncontextualized setting.

We adopt the
setup in \citet{yaghoobzadeh-etal-2019-probing} to probe
uncontextualized embeddings --
for each of the 34 s-classes, we train
a binary classifier as shown in
\figref{typeleveltrainer}.
\Tabref{datastas}, column
\emph{words} shows the sizes of
train/dev/test.
The evaluation measure is micro $F_1$
over all decisions of the 34 binary
classifiers.

\begin{figure}
  \vspace{-0.4cm}
  \begin{algorithm}[H]
    \small
    \algrenewcommand\algorithmicindent{0.5cm}
    \caption{\small Train a classifier with type-level embeddings}
    \begin{algorithmic}[1]
      \Procedure{TypeSclsTrainer}
      {Dict: word2vec, Dict: word2sclass, sclass: $\mathcal{S}$, List: TrainWords}:
      \State PosVecs, NegVecs = [], []
      \For {word $\in$ TrainWords}
      \State vector = word2vec.get(word)
      \State sclasses = word2sclass.get(word)
      \If {$\mathcal{S} \in$ sclasses}
      \State PosVecs.append(vector)
      \Else
      \State NegVecs.append(vector)
      \EndIf
      \EndFor
      \State classifier = Classifier()
      \State classifier.train(PosVecs, NegVecs)
      \State \Return classifier
      \EndProcedure
    \end{algorithmic}
  \end{algorithm}
  \vspace{-0.5cm}
  \caption{Training a diagnostic classifier with uncontextualized word
    representations for an s-class $\mathcal{S}$.}
  \figlabel{typeleveltrainer}
\end{figure}

\subsubsection{Probing contextualized embeddings}
\seclabel{probctxemb}
We probe BERT with the same setup:
a binary classifier is trained for each of the 34 s-classes;
each BERT layer is probed individually.

For uncontextualized embeddings, a word has a  single
vector, which is either a
positive or negative example for an s-class. For contextualized embeddings,
the contexts of a word will typically be mixed; for example, ``food'' contexts
(a candy) of ``@airheads@'' are positive but ``art'' contexts (a film) of
``@airheads@'' are negative examples for the classifier of
``food''. \Tabref{datastas}, column
\emph{contexts} shows the sizes of train/dev/test when probing BERT.
\figref{probingmodel} compares our two probing setups.

In evaluation, we weight frequent word-s-class combinations (those
having 100 contexts in our dataset) and the much larger number of
less frequent word-s-class combinations equally. To this end,
we aggregate the decisions for the contexts of a word-s-class combination. We
stipulate that at least half of the contexts must be correctly classified.
For example, ``@airheads@-art'' occurs 47 times, so we evaluate the
``art'' classifier as accurate for ``@airheads@-art'' if it classifies
at least 24 contexts correctly. The final evaluation
measure is micro $F_1$ over
all 15,437 (for dev) and 77,706 (for test) decisions (see
\tabref{datastas}) of the 34 classifiers for the word-s-class combinations.

\begin{figure}
  \centering
  \includegraphics[width=.8\linewidth]{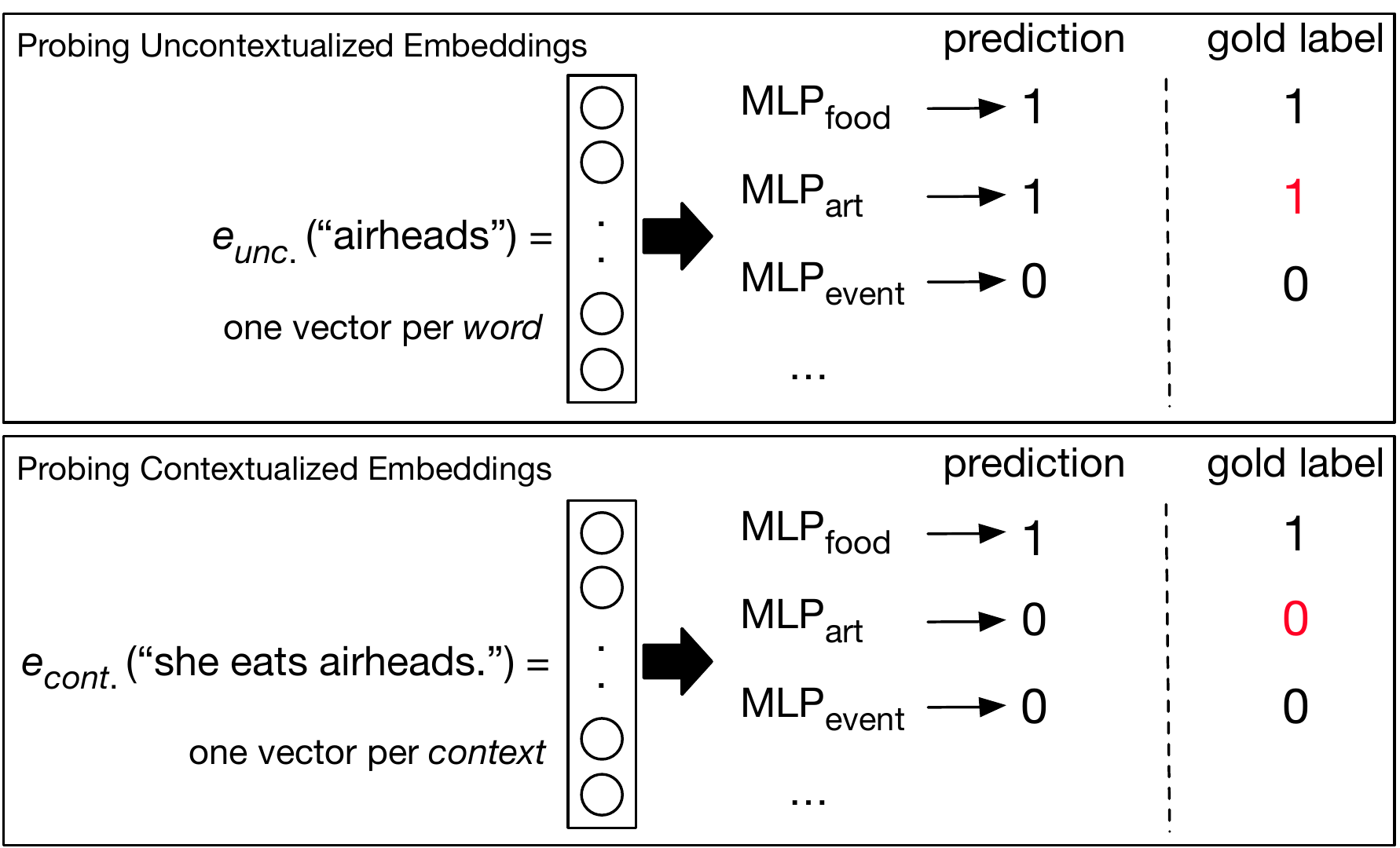}
  \caption{Setups for probing uncontextualized and contextualized embeddings.
    For BERT, we input a context sentence to
    extract the contextualized embedding of a word, e.g.,
    ``airheads''; ``food'' is the correct s-class label for this context.}
  \figlabel{probingmodel}
\end{figure}

\section{Experiments and Results}
\subsection{Data preprocessing}
BERT uses wordpieces  \citep{wu2016google} to
represent text and infrequent words are tokenized to several
wordpieces. For example, ``infrequent'' is tokenized to ``in'', ``\#\#fr'',
``\#\#e'', and ``\#\#quent''. Following \citet{he2019establishing}, we average wordpiece
embeddings to get a single vector representation of a
word.\footnote{Some ``words'' in Wiki-PSE are in reality
  multiword phrases. Again, we average in these cases
  to get a single vector representation.}

We limit the maximum sequence length of the context
sentence input to BERT to 128.
Consistent with the probing literature, we use a simple
probing classifier: a 1-layer multilayer perceptron (MLP) with 1024 hidden
dimensions and ReLU.

\subsection{Quantifying contextualization}
\seclabel{fullexp}
\subsubsection{Representation learners}
\seclabel{representationlearners}
Six \textbf{uncontextualized embedding spaces} are evaluated:
(i) PSE. A 300-dimensional embedding space computed by running
skipgram with negative sampling \citep{mikolov2013distributed} on the
Wiki-PSE text corpus. \citet{yaghoobzadeh-etal-2019-probing} show that PSE
outperforms other  embedding spaces.
(ii) Rand. An embedding space with the same vocabulary and dimension size as PSE.
Vectors are drawn from $\mathcal{N} (\textbf{0}, \textbf{I}_{300})$.
Rand is used to confirm that word
representations indeed encode valid meaning contents that can be identified
by diagnostic MLPs rather than random weights.
(iii) The 300-dimensional fastText \citep{bojanowski-etal-2017-enriching} embeddings.
(iv) GloVe. The
300-dimensional space trained on 6 billion tokens \citep{Pennington2014}.
Out-of-vocabulary (OOV) words are associated with vectors drawn from $\mathcal{N} (\textbf{0},
  \textbf{I}_{300})$.
(v) BERTw. The 768-dimensional wordpiece
embeddings in BERT. We tokenize a word
with the BERT tokenizer then average its wordpiece embeddings.
(vi) AVG-BERT-$\ell$.\footnote{BERTw and AVG-BERT-$\ell$ have more dimensions. But
  \citet{yaghoobzadeh-etal-2019-probing} showed that
  different dimensionalities have a negligible impact on
  relative performance when
  probing for s-classes using MLPs as diagnostic classifiers.}
For an annotated word in Wiki-PSE,
we average all of its contextualized embeddings from BERT
layer $\ell$ in the Wiki-PSE text corpus.
Comparing AVG-BERT-$\ell$ with others brings a new insight:
to which extent does this ``uncontextualized'' variant of BERT outperform
others in encoding different s-classes of a word?

Four \textbf{contextualized embedding models} are considered:
(i) BERT. We use the PyTorch \citep{paszke2017automatic, wolf2019transformers}
implementation of the 12-layer BERT-base-uncased
model (Wiki-PSE is uncased).
(ii) P-BERT. A bag-of-word model that ``contextualizes''
the wordpiece embedding of an annotated word by
averaging the embeddings of wordpieces of the sentence
it occurs in.
Comparing BERT with P-BERT reveals to which extent the self attention
mechanism outperforms an average pooling practice when contextualizing words.
(iii) P-fastText. Similar to P-BERT, but we use
fastText word embeddings.
Comparing BERT with P-fastText indicates
to which extent BERT outperforms
uncontextualized embedding spaces when they also have access to contextual information.
(iv) P-Rand. Similar to P-BERT, but we draw word embeddings from
$\mathcal{N}(\textbf{0}, \textbf{I}_{300})$.
\citet{wieting2019no} show that a random baseline
has good performance in tasks like sentence classification.

\begin{figure}
  \centering
  \includegraphics[width=.7\linewidth]{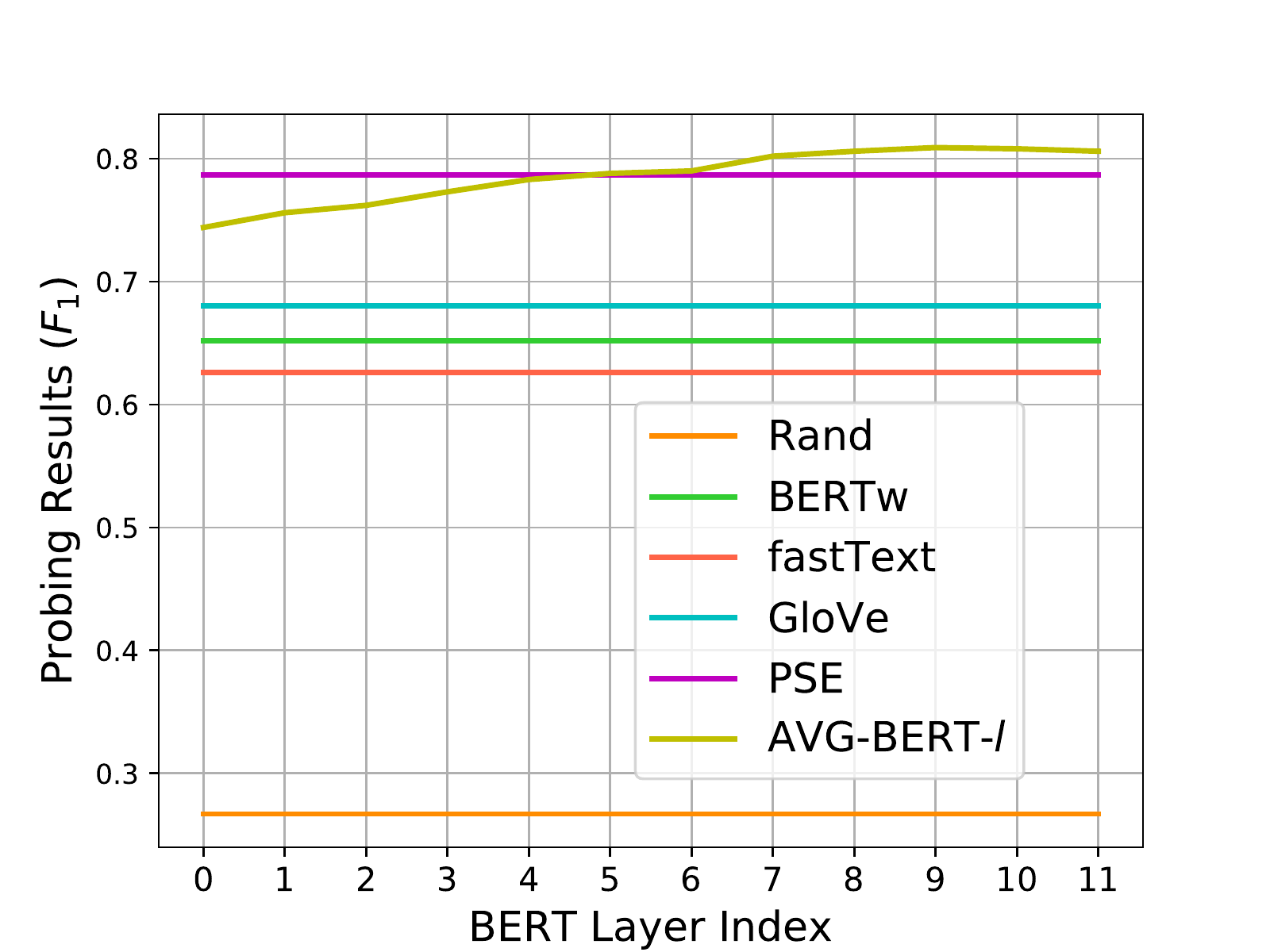}
  \caption{S-class probing results
  for \textbf{uncontextualized} embeddings.
  Results are micro $F_1$ on Wiki-PSE test set.
  Numerical values are in \tabref{uncontextualizedprob} in Appendix.}
  \figlabel{unctxprobres}
\end{figure}

\subsubsection{S-class inference results}
\seclabel{basicexp}
\figref{unctxprobres} shows
\textbf{uncontextualized embedding}
probing results.
Comparing with random weights, all embedding spaces encode informative features
helping s-class inference. BERTw delivers
results similar to GloVe and fastText, demonstrating our earlier point (cf.\ the
qualitative example in \tabref{suitexample}) that the lowest embedding layer of BERT
is uncontextualized;
several meanings of a word are conflated into a single vector.

PSE performs strongly, consistent with observations in
\citet{yaghoobzadeh-etal-2019-probing}.
AVG-BERT-10 performs best among all spaces. Thus for a
given word, averaging its contextualized embeddings from BERT yields a high quality
type-level embedding vector, similar to ``anchor words'' in cross-lingual alignment
\citep{schuster-etal-2019-cross}.

As expected, the top AVG-BERT layers outperform lower layers, given
the deep architecture of BERT. Additionally, AVG-BERT-0 significantly
outperforms BERTw, evidencing the importance of position embeddings
and
the self attention mechanism \citep{vaswani2017attention} when composing the
wordpieces of a word.

\begin{figure}
  \centering
  \includegraphics[width=.7\linewidth]{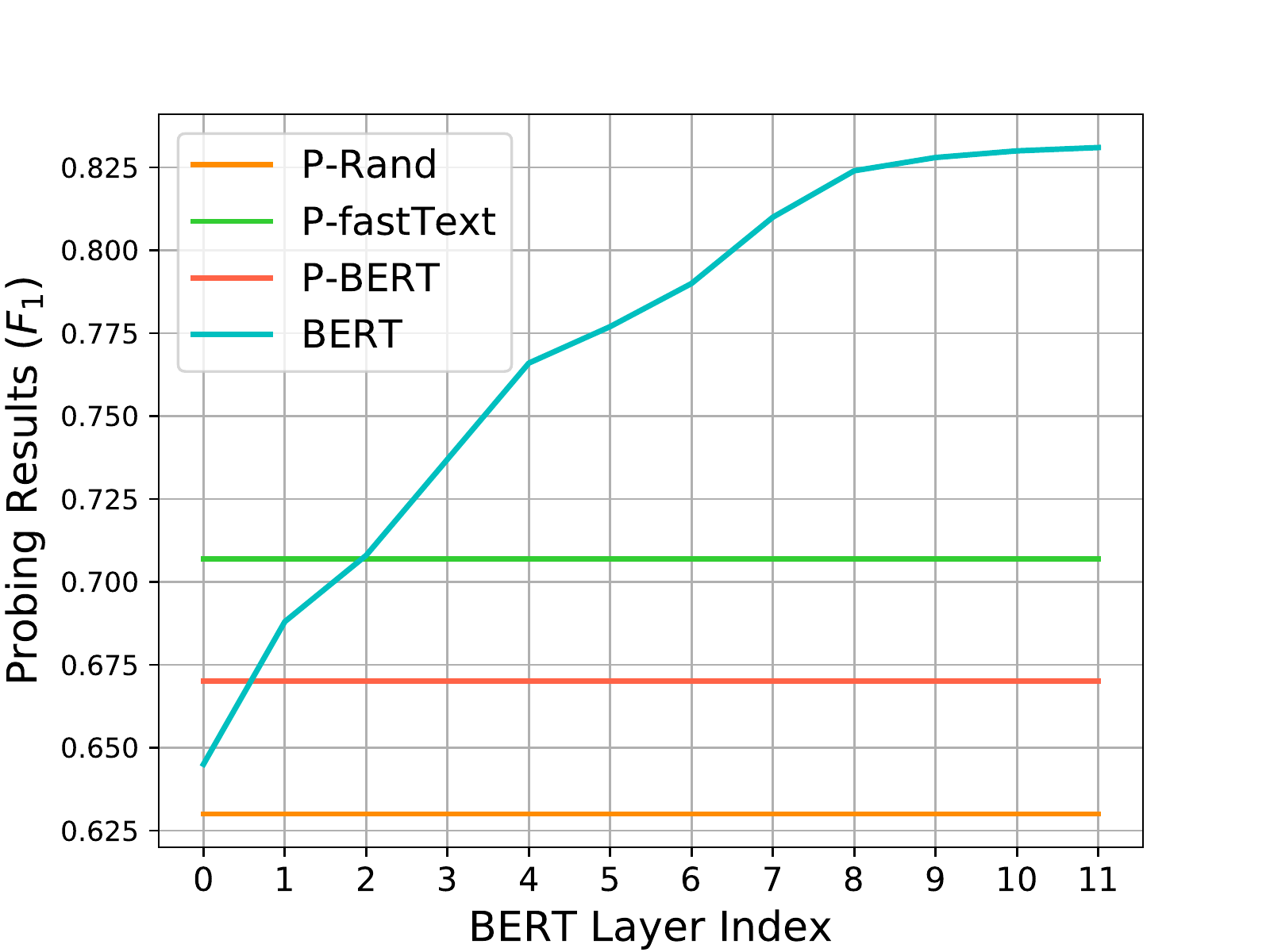}
  \caption{S-class probing results for \textbf{contextualized} embedding models.
    Results are micro $F_1$ on Wiki-PSE test set.
    Numerical values are in \tabref{contextualizedprob} in  Appendix.}
  \figlabel{ctxprobres}
\end{figure}

\figref{ctxprobres} shows
\textbf{contextualized embedding} probing results.
Comparing BERT layers, a clear trend can be identified: s-class
inference performance increases monotonically with higher layers. This
increase levels off in the top layers. Thus,
the features from deeper layers improve word contextualization,
advancing s-class inference. 
It also verifies previous findings: semantic
tasks are mainly solved at higher layers \citep{liu-etal-2019-linguistic,
  tenney-etal-2019-bert}.
We can also observe that the strongest contextualization occurs early at
lower layers --
going up to layer 1 from layer 0 brings a 4\% (absolute) improvement.

The very limited contextualization improvement brought by the top two layers
may explain why representations from the top layers of
BERT can deliver
suboptimal performance on  NLP tasks \citep{liu-etal-2019-linguistic}:
the top layers are optimized for the pretraining objective, i.e., predicting
masked words \citep{voita-etal-2019-bottom}, not for the contextualization
of words that is helpful for NLP tasks.

BERT layer 0 performs slightly worse than P-BERT, which
may be due to the
fact that some attention heads in lower layers of BERT attend broadly
in the sentence, producing ``bag-of-vector-like'' representations
\citep{clark-etal-2019-bert},
which is in fact close to the setup of P-BERT.
However, starting from layer 1, BERT gradually improves  and
surpasses P-BERT,
achieving a maximum gain of
0.16
in $F_1$ in layer 11.
Thus, BERT knows how to better
interpret the word in context, i.e., contextualize
the word, when progressively going to deeper (higher) layers.

P-Rand performs strongly, but is noticeably worse than P-fastText and P-BERT.
P-fastText outperforms P-BERT and BERT layers 0 and 1.
We hypothesize that 
this may be due to the fact that fastText learns embeddings directly for words;
P-BERT and BERT have to compose subwords to understand the meaning of a word, which
is more challenging. Starting from layer 2, BERT outperforms
P-fastText and P-BERT, illustrating the effectiveness of self attention in
better integrating the information from the context
into contextualized word embeddings than the average pooling practice
in bag-of-word models.

\figref{unctxprobres} and \figref{ctxprobres} jointly illustrate the high quality
of word representations computed by BERT. The
BERT-derived uncontextualized
AVG-BERT-$\ell$ representations -- modeled as
\citet{schuster-etal-2019-cross}'s anchor words --
show superior capability in inferring s-classes of a word,
performing best among all uncontextualized embeddings.
This suggests that
BERT's powerful
learning architecture
may be the main reason for BERT's high performance, not
contextualization proper, i.e.,
the representation of words  as contextualized embeddings
on the highest layer when BERT is applied to NLP tasks.
This offers intriguing possibility for creating (or distilling)
strongly
performing uncontextualized BERT-derived
models that are more compact and more efficiently deployable.

\subsubsection{Qualitative analysis}
\secref{basicexp} quantitatively shows that BERT performs strongly in
contextualizing words, thanks to its deep integration of information from the
entire input sentence in each contextualized embedding. But there are
scenarios where BERT fails. We identify two such cases in which the
contextual information does not help s-class inference.

(i) \textbf{Tokenization}. In some domains,  the annotated word  and/or its
context words are tokenized into several wordpieces due to their low frequency
in the pretraining corpora.
As a result, BERT may not
be able to derive the correct composed meaning.  Then the
MLPs cannot identify the correct s-class from
the noisy input. Consider the tokenized results of ``@glutamate@-biology''
and one of its contexts:

\emph{``three ne \#\#uro \#\#tra \#\#ns \#\#mit \#\#ters that play
  important roles in adolescent brain development are \textbf{g \#\#lu
    \#\#tama \#\#te} \ldots''}

Though ``brain development'' hints at a context related to ``biology'', this
signal could be swamped by the noise in embeddings of other -- especially
short -- wordpieces. 
\citet{schick2019rare} propose
a mimicking approach
\citep{pinter-etal-2017-mimicking} to help BERT 
understand rare words.

(ii) \textbf{Uninformative contexts}. Some contexts do not provide sufficient information
related to the s-class. For example, according to probing results on BERTw,
the wordpiece embedding of ``goodfellas'' does not encode the meaning of
s-class ``art'' (i.e., movies); the context ``Chase also said he wanted Imperioli
because he had been in Goodfellas'' of word-s-class combination ``@goodfellas@-art'' is not
informative enough for inferring an ``art'' context, yielding incorrect
predictions in higher layers.

\subsection{Context size}
\seclabel{ctxlvl}
We now quantify the amount of context required by BERT for 
properly contextualizing words to produce accurate s-class inference results.

When probing for the s-class of word $w$,
we define \emph{context size} as the number of words
surrounding $w$ (left and right) in a sentence before wordpiece tokenization.
For example, a
context size of 5 means 5 words left, 5 words right.
The context size seems to be picked heuristically in other work.
\citet{yarowsky-1992-word} and \citet{gale1992one} use 50 while
\citet{black1988experiment} uses 3--6. We experiment with a range of
context sizes then compare s-class inference results.
We also enclose P-BERT for comparison.
Note that this experiment is different from edge probing
\citep{tenney2019you}, which takes the full sentence as
input. We only make input words within the context window available
to BERT and P-BERT.

\begin{figure}
  \centering
  \includegraphics[width=.8\linewidth]{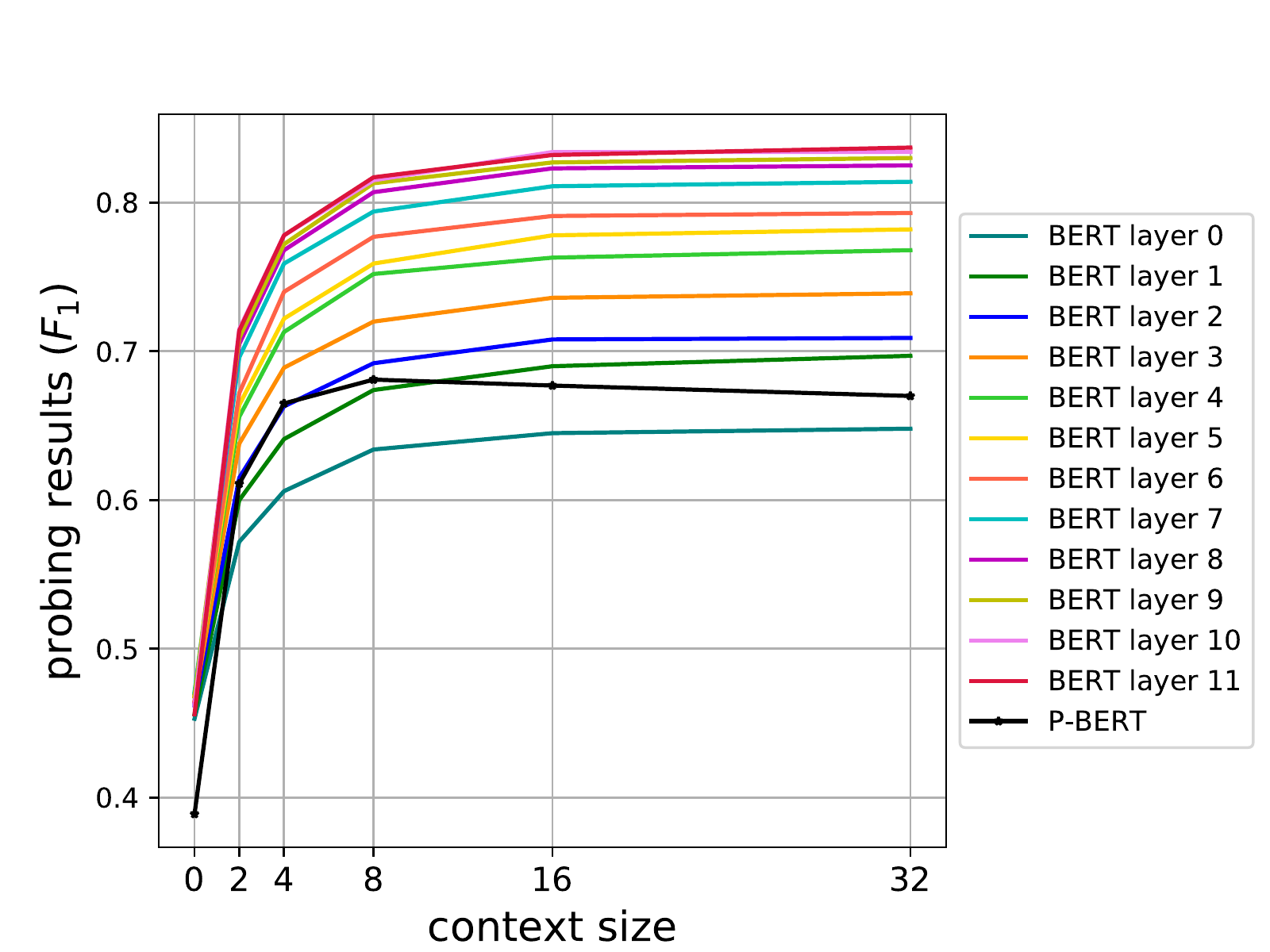}
  \caption{Probing results on the dev set with
    different context sizes.
    For BERT, performance increases with context size.
    Large context sizes like 16 and 32 slightly hurt performance of P-BERT.}
  \figlabel{contextlevel}
\end{figure}

\subsubsection{Probing results}
We report micro $F_1$ on Wiki-PSE dev, with context size $\in \{0, 2, 4, 8, 16,
  32\}$. Context size 0 means that the input consists only of the wordpiece
  embeddings of the input word. \figref{contextlevel} shows results.

\textbf{Comparing context sizes}. Larger context sizes have higher performance
for all BERT layers. Improvements are most prominent for small context sizes,
e.g., 2 and 4, meaning that often local features are sufficient to 
contextualize words and infer
s-classes, supporting  \citet{black1988experiment}'s
design choice of
3--6. 
Further increasing the context size improves contextualization only marginally.

A qualitative example showing informative local features
is ``The Azande speak Zande, which they call Pa-Zande.''
In this context, the gold s-class of ``Zande'' is ``language''
(instead of ``people-ethnicity'', i.e., the Zande people).
The MLPs for BERTw and for
context size 0 for BERT fail to identify
s-class ``language''.
But the  BERT MLP for context size 2 predicts ``language'' correctly
since it includes the strong signal ``speak''. This
context is a case of
selectional restrictions
\citep{resnik-1993-semantic,Jurafsky:2009:SLP:1214993}, in
this case possible objects of
``speak''.

As small context sizes already contain noticeable
information contextualizing the words,
we hypothesize that it may not be necessary
to exploit the full context in cases
where the
quadratic complexity of full-sentence self attention is problematic, e.g., on edge devices.
Initial results on part-of-speech tagging with the Penn Treebank \citep{marcus1993building}
in Appendix \secref{ctxpos} confirm our hypothesis.
We leave more experiments to future work.

P-BERT shows a similar pattern when varying the context sizes.
However, large context sizes such as 16 and 32 hurt contextualization,
meaning that averaging too many embeddings
results in a bag of words not specific to a particular token.

\textbf{Comparing BERT layers}.
Higher layers of BERT yield better contextualized word embeddings.
This phenomenon is more noticeable for large context sizes
such as 8, 16 and 32. However for small context sizes,
e.g., 0, embeddings
from all layers perform similarly and badly. This means that without context
information, simply passing the wordpiece embedding of a word through
BERT layers does not help, suggesting that contextualization is the key
ability of BERT yielding impressive performance across NLP tasks.

Again, P-BERT only outperforms layer 0 of BERT with most context sizes,
suggesting that BERT layers, especially the top layers, contextualize words
with abstract and informative representations, instead of naively
aggregating all information within the context sentence.

\subsection{Probing finetuned embeddings}
\seclabel{reprob}
We have done ``classical'' probing: extracting features from
pretrained BERT and feeding them to diagnostic classifiers. 
However, pretrained BERT needs to be adapted, i.e., finetuned, for
good performance on tasks \citep{devlin-etal-2019-bert, peters-etal-2019-tune}.
Thus, it is necessary to 
investigate how finetuning BERT affects the contextualization of words
and analyze how the pretrained knowledge and probed features change.

\renewcommand\thesubfigure{\roman{subfigure}}
\begin{figure*}
  \centering
  \begin{subfigure}{.5\textwidth}
    \centering
    \includegraphics[scale=0.4]{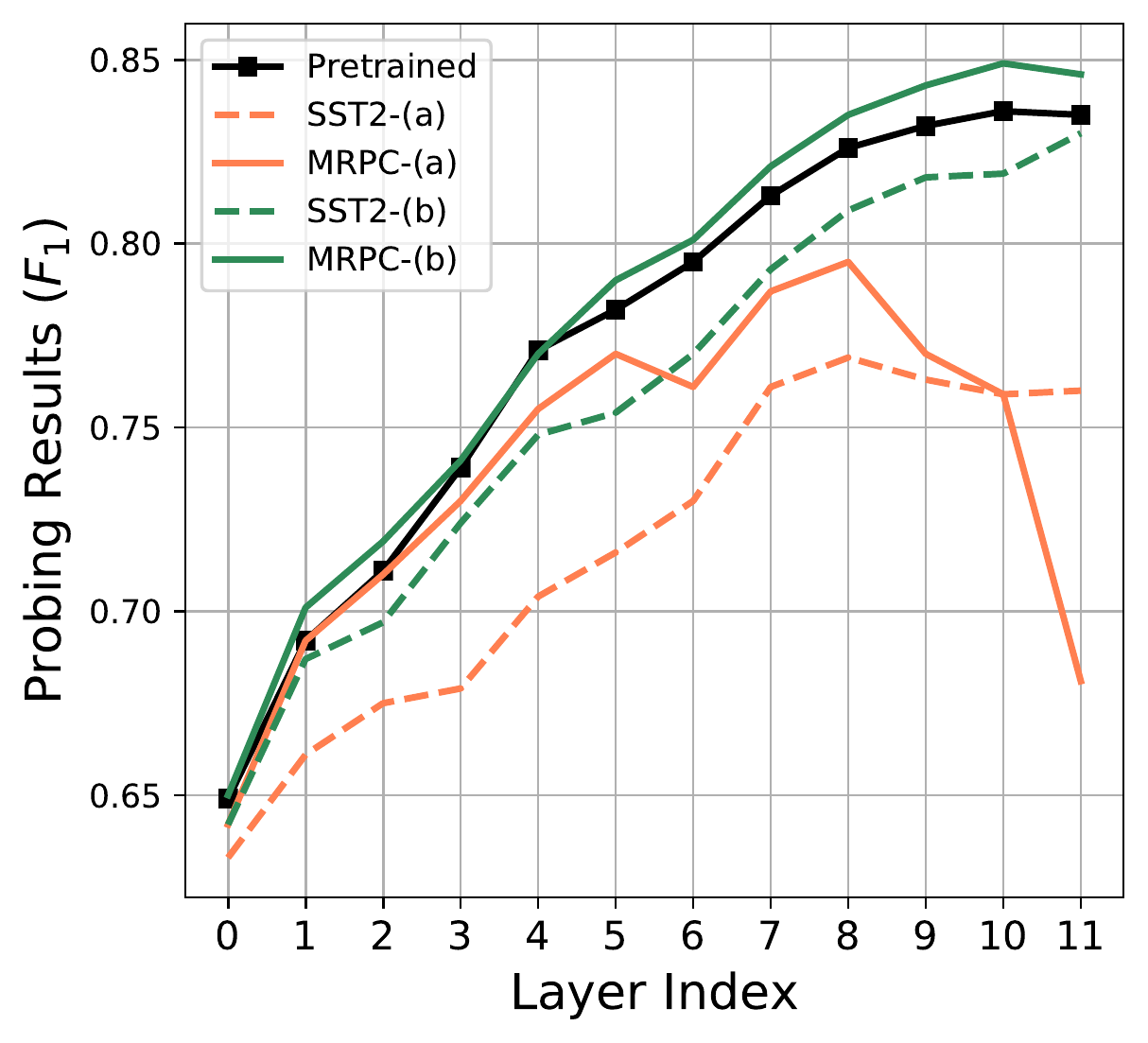}
    \caption{MRPC and SST2}
  \end{subfigure}%
  \begin{subfigure}{.5\textwidth}
    \centering
    \includegraphics[scale=.4]{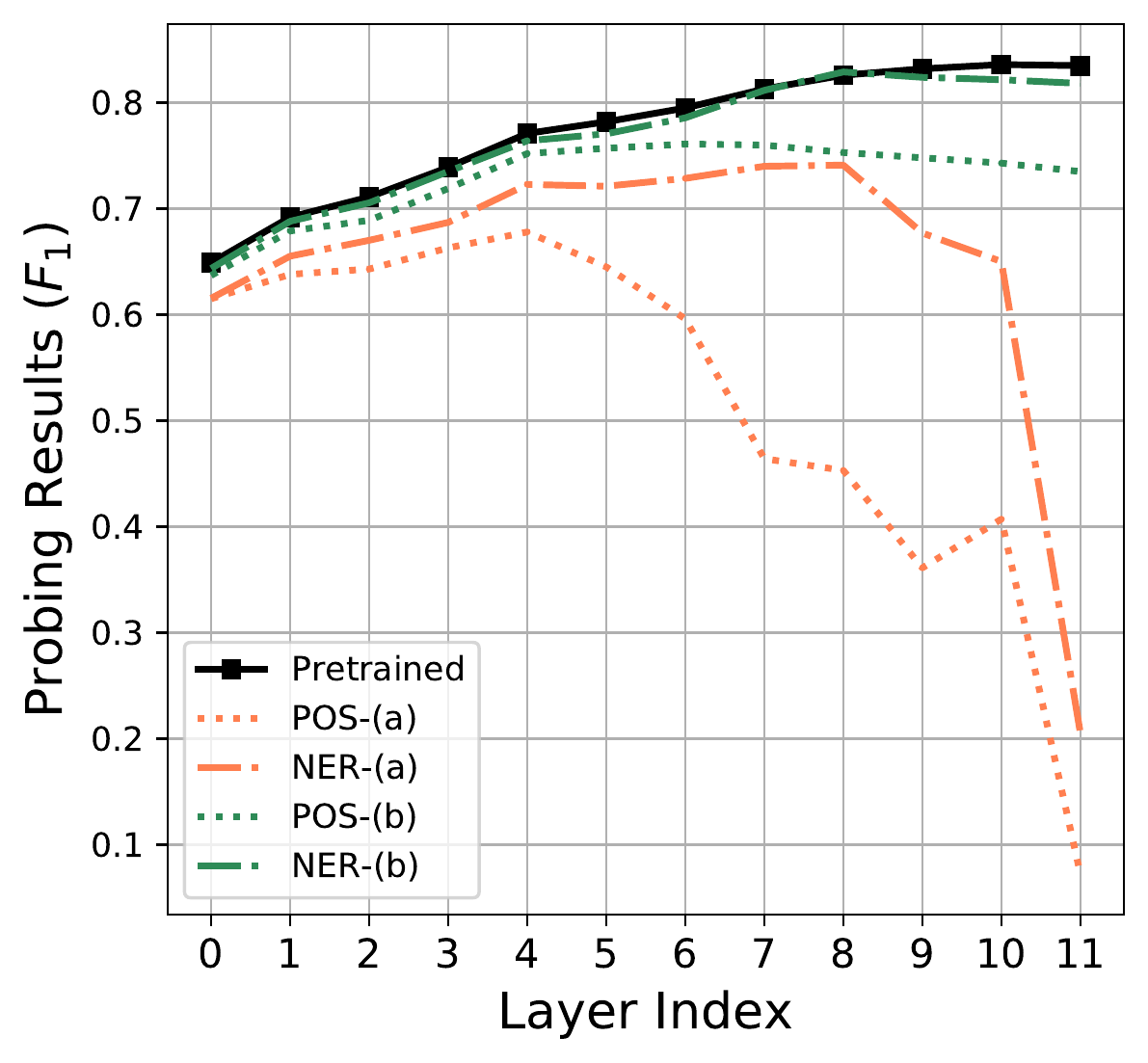}
    \caption{POS and NER}
  \end{subfigure}
  \caption{
    Comparing s-class inference results of pretrained BERT and BERT
    finetuned on MRPC, SST2, POS, and NER.  ``Pretrained'': probing
    results on weight-frozen pretrained BERT in \secref{fullexp}.
    For (a), we directly apply the MLPs in \secref{fullexp} (trained
    with pretrained embeddings) to finetuned BERT embeddings; for (b),
    we train and evaluate a new set of MLPs on the finetuned BERT
    embeddings.  } \figlabel{reprobe}
\end{figure*}

\subsubsection{Finetuning tasks}
We finetune BERT on four tasks:  
part-of-speech (POS) tagging on the Penn Treebank \citep{marcus1993building}, 
named-entity recognition (NER) on the CoNLL-2003 Shared Task \citep{nerconll2003},
binary sentiment classification on the Stanford Sentiment Treebank  (SST2)
\citep{socher2013recursive}
and paraphrase detection on the Microsoft Research
Paraphrase Corpus (MRPC) \citep{dolan2005automatically}. 
For SST2 and MRPC, we
use the GLUE train and dev sets  \citep{wang2018glue}. For POS,
sections 0-18 of WSJ are train and sections 19-21 are
dev  \citep{collins-2002-discriminative}.
For NER, we use the official data splits.

Following \citet{devlin-etal-2019-bert}, we put a linear layer
on top of the pretrained BERT, then finetune all parameters.
We use Adam \citep{kingma2014adam}
with learning rate 5e-5 for 5 epochs. We
save the model from the step that performs best on dev
(of MRPC/SST2/POS/NER), extract representations from Wiki-PSE
using this model and then report results on Wiki-PSE dev.

\begin{table}[t]
  \renewcommand{\arraystretch}{1.2}
  \centering\small
  \begin{tabular}{r|rrrr}
           & POS  & SST2 & MRPC & NER\\  \hline
  Ours     & .977 & .928  & .853 & .946 \\
  \citet{devlin-etal-2019-bert}    & n/a  & .927  & .867 & .964
  \end{tabular}
  \caption{Dev set performance of finetuning BERT (bert-base-uncased). 
  For NER, we report micro $F_1$. For other tasks, we report accuracy.}
\label{finetuneres}
\end{table}

Table \ref{finetuneres} reports the finetuning results.  Our finetuned
models perform comparably to \citet{devlin-etal-2019-bert} on SST2 and
MRPC. Our NER result is slightly worse, this may due to the fact that
\citet{devlin-etal-2019-bert} use ``maximal document
context''
while we use sentence-level
context of 128 max sequence length.
More finetuning details are available in Appendix \secref{finetuningdetails}.

\subsubsection{Probing results}
We now quantify the contextualization of word representations from
finetuned BERT models. 
Two setups are considered:
(a) directly
apply the MLPs in \secref{fullexp} (trained with pretrained embeddings) to
finetuned BERT embeddings;
(b) train and evaluate a new set of MLPs on the finetuned BERT embeddings.

Comparing (a) with probing results on pretrained BERT (\secref{fullexp})
gives us an intuition about how many changes occurred to the knowledge captured
during pretraining. Comparing (b) with \secref{fullexp} reveals whether or not the
pretrained knowledge about contextualization is still preserved in finetuned models.

\figref{reprobe} shows s-class probing results of finetuned BERT with setup (a) and (b).
For example in (ii), layer 11 s-class inference performance of the POS-finetuned BERT
decreases by 0.763 (0.835 $\rightarrow$ 0.072, from ``Pretrained'' to ``POS-(a)'')
when using the MLPs from \secref{fullexp}.

\textbf{Comparing setup (a) and ``Pretrained''},
we see that finetuning brings significant changes to the word representations.
Finetuning on POS and NER 
introduces more obvious probing accuracy drops
than finetuning on SST2 and MRPC.
This may be due to the fact that the training objective of SST2 and MRPC
takes as input only the \texttt{[CLS]} token
while all words in a sentence 
are involved in the training objective of POS and NER.

\textbf{Comparing setup (b) and ``Pretrained''}. 
Finetuning BERT on MRPC introduces small
but consistent improvements on s-class inference. 
For SST2 and NER, very small s-class inference accuracy drops are observed.
Finetuning on POS brings more noticeable changes.
Solving POS requires more syntactic information than the other tasks,
inducing BERT to
``propagate'' the syntactic information that is represented
in lower layers to the upper layers;
due to
their limited capacity,
the fixed-size vectors from the upper layers may lose some
semantic information,
yielding a more noticeable performance drop on s-class
inference.

\textbf{Comparing (a) and (b)},
we see that the knowledge about contextualizing words captured
during pretraining
is still well preserved after finetuning. For example,
the MLPs trained with layer 11 embeddings computed by the
POS-finetuned BERT still
achieve a reasonably good score of 0.735 (a 0.100 drop
compared with ``Pretrained'' -- compare black and green
dotted lines in  \figref{reprobe} (ii)).
Thus, the semantic
information needed for inferring s-classes is still present
to a large extent.

Finetuning may introduce large changes (setup (a))
to the representations
--
similar to the projection utilized to uncover
divergent information in uncontextualized word embeddings
\citep{artetxe-etal-2018-uncovering} --
but relatively little
information about contextualization is lost
as the good performance of the newly
trained MLPs shows (setup (b)). 
Similarly, \citet{merchant2020happens} show that finetuned BERT
still well preserves the probed ``linguistic features'' in pretrained BERT.

\textbf{Comparing BERT layers}.
Contextualized embeddings
from BERT's top layers
are strongly affected by finetuning, especially for setup (a).
In contrast, lower
layers are more invariant and show s-class
inference results similar to the pretrained model.
\citet{hao2019visualizing}, \citet{lee2019elsa}, \citet{kovaleva2019revealing}
make similar observations:
lower layer representations
are more
transferable across different tasks and top layer representations
are more task-specific after finetuning.

\figref{heatmap} shows
the cosine similarity of the flattened self attention
weights computed by pretrained, POS-, and MRPC-finetuned BERT
using the dev set examples.
We see that
top layers are more sensitive to
finetuning (darker color) while lower layers are barely
changed (lighter color).
Top layers have more changes for POS than for MRPC,
in line with probing results in
\figref{reprobe}.

\renewcommand\thesubfigure{\roman{subfigure}}
\begin{figure}
  \centering
  \begin{subfigure}{.25\textwidth}
    \centering
    \includegraphics[scale=0.25]{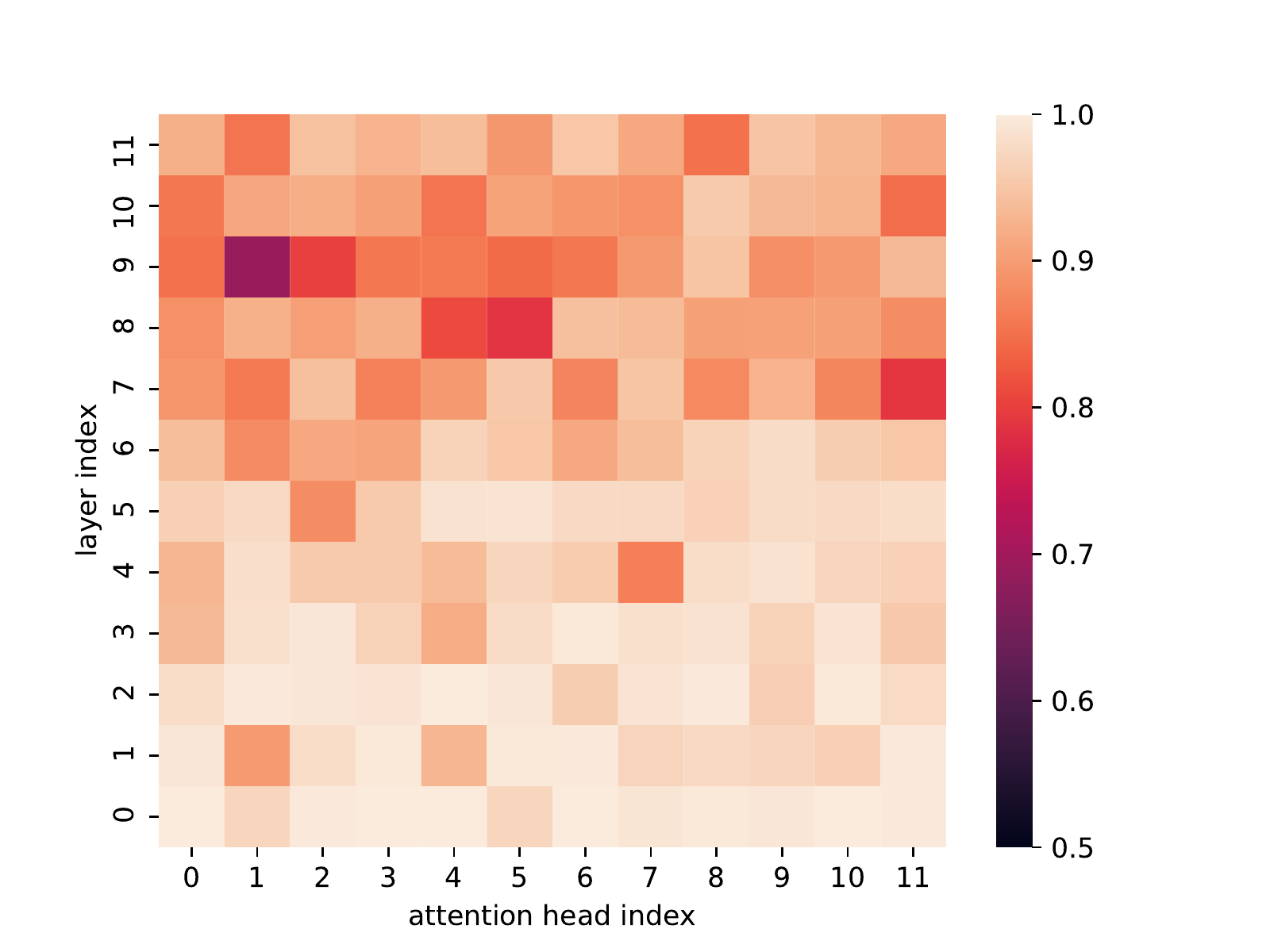}
    \caption{pretrained vs.\ POS}
  \end{subfigure}%
  \begin{subfigure}{.25\textwidth}
    \centering
    \includegraphics[scale=.25]{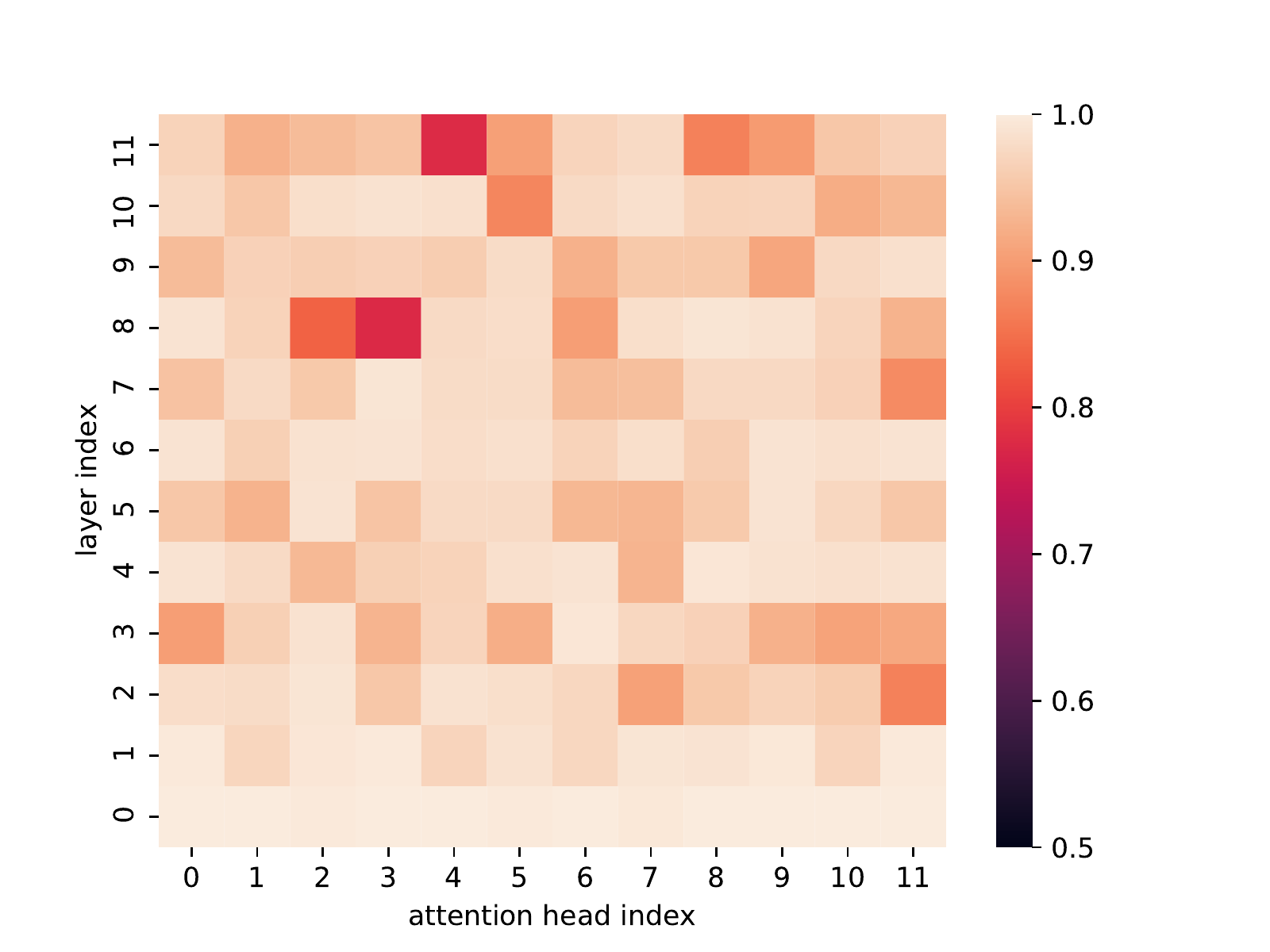}
    \caption{pretrained vs.\ MRPC}
  \end{subfigure}
  \caption{Cosine similarity of flattened self attention weights. 
  X-axis: index of the 12 self attention heads; 
  y-axis: layer index. Darker colors: smaller
    similarities, i.e., larger changes brought by finetuning.}
  \figlabel{heatmap}
\end{figure}

\section{Related Work}
\textbf{Interpreting deep  networks}.
\Ptls \citep{mccann2017learned, peters-etal-2018-deep, radford2019language,
  devlin-etal-2019-bert}  advance NLP by
contextualized representations of words.
A key goal of current research is to understand how these
models work and what they represent on different layers.

Probing is a recent strand of work that
investigates -- via diagnostic classifiers --
desired syntactic and semantic features  encoded in \ptl  representations.
\citet{shi-etal-2016-string} show that
string-based RNNs  encode syntactic information.
\citet{belinkov-etal-2017-evaluating} investigate word representations at
different layers in NMT.
\citet{linzen-etal-2016-assessing}
assess the syntactic ability of LSTM
\citep{hochreiter1997long} encoders
and  \citet{goldberg2019assessing}
of BERT.
\citet{tenney-etal-2019-bert} find that information on
POS tagging, parsing, NER, semantic roles, and
coreference is represented on increasingly higher layers
of BERT.
\citet{yaghoobzadeh-etal-2019-probing} assess the disambiguation properties of
type-level word
representations. \citet{liu-etal-2019-linguistic} and \citet{lin2019open}
investigate the linguistic knowledge encoded in BERT.
\citet{adi2016fine}, \citet{conneau-etal-2018-cram}, and \citet{wieting2019no}
study sentence embedding properties via probing.
\citet{peters-etal-2018-dissecting} probe how the network architecture affects
the learned vectors.

In all of these studies, probing
serves to  analyze representations and reveal their properties.
We employ probing to investigate the
contextualization of words in \ptls quantitatively.
In addition, we exploit how finetuning affects word contextualization.

\citet{ethayarajh2019contextual} quantitatively investigates
contextualized embeddings, 
using unsupervised cosine-similarity-based evaluation.
Inferring s-classes, we address a complementary set of
questions because we can
quantify contextualization
with a
uniform set of semantic classes.
\citet{Brunner2020On} employ token identifiability to compute the
deviation of a contextualized embedding from the uncontextualized
embedding.
\citet{voita-etal-2019-bottom} address this from the mutual
information perspective, e.g., low mutual information between an
uncontextualized embedding and its contextualized embedding can be
viewed as a reflection of more contextualization.  Similar
observations are made: higher layer embeddings are more contextualized
while lower layer embeddings are less contextualized.
In contrast, we draw the observations from the perspective of s-class
inference.  The higher layer embeddings perform better when evaluating
the semantic classes -- they are better contextualized and have higher
fitness to the context than the lower layer embeddings.

\textbf{Two-stage NLP paradigm}. 
Recent work \citep{dai2015semisupervised,howard-ruder-2018-universal,
  devlin-etal-2019-bert}
introduces a ``two-stage paradigm'' in NLP:
pretrain a language encoder on a large amount of unlabeled data via
self-supervised learning, then finetune the encoder on task-specific
benchmarks like GLUE \citep{wang2018glue, wang2019superglue}. This
transfer-learning pipeline yields good and robust results
compared to models
trained from scratch \citep{hao2019visualizing}.

In this work, we shed light on how BERT's pretrained knowledge
about contextualization changes during finetuning
by comparing s-class inference ability of pretrained and
finetuned models.
\citet{merchant2020happens} analyze BERT models finetuned on different
downstream tasks with the edge probing suite \citep{tenney2019you} and
make similar observations as us. They focus on ``linguistic features''
while we focus on the contextualization of words.

\section{Conclusion}
We presented a quantitative study of the contextualization
of words in BERT by investigating BERT's semantic class
inference capabilities.  
We focused on two key
factors for successful contextualization by BERT: layer
index and context size.
By comparing pretrained and finetuned models, 
we showed that word-level tasks like part-of-speech tagging 
bring more noticeable changes than
sentence-level tasks like paraphrase classification;
and top layers of BERT are more
sensitive to the finetuning objective than lower layers.
We also found that BERT's pretrained knowledge about contextualizing words
is still well retained
after finetuning.

We showed that exploiting the full context may be unnecessary in
applications where the quadratic complexity of full-sentence attention
is problematic.  Future work may evaluate this phenomenon on more datasets
and downstream tasks.

\section*{Acknowledgments}
We thank
the anonymous reviewers for the insightful comments and suggestions.
This work was funded by the European Research Council (ERC \#740516)
and a Zentrum Digitalisierung.Bayern fellowship award.

\bibliography{emnlp2020}
\bibliographystyle{acl_natbib}

\clearpage
\appendix
\input{appendix}

\end{document}

%% file: appendix.tex
  \section{Reproducibility Checklist}

  \subsection{Computing infrastructure} 
  All experiments are conducted on GeForce GTX 1080 Ti and GeForce GTX 1080. 

  \subsection{Number of parameters}
  We use a set of 34 binary MLPs to conduct our probing task. Each MLP has input dimension 768,
  hidden dimension 1024 and output dimension 2. As a result, the total number of parameters
  is 26,843,204.
  For finetuning, we use the BERT-base-uncased model containing about
  110 million parameters
  (\url{https://github.com/google-research/bert}).

  \subsection{Validation performance}
  Following \tabref{uncontextualizedprob} and \tabref{contextualizedprob} report the 
  validation performance of probing uncontextualized and contextualized embeddings.

  \subsection{Evaluation metric}
  Our evaluation is the micro $F_1$ over all decisions of the 34 probing
  classifiers. More details are available in \secref{probingexperiments} of the main paper.

  \subsection{Hyperparameter search}
  For probing tasks, we do not conduct hyperparameter search since our
  goal is to analyze the contextualization. The probing classifiers are trained
  with learning rate 1e-3 and 400 epochs.
  For finetuning BERT, we do not search hyperparameters but directly adopt the
  setup in \citet{devlin-etal-2019-bert} as shown in Table \ref{hypers}.

  \subsection{Datasets}
    \seclabel{psedetails}
    List of the 34 semantic classes (s-classes), 
  number of word-s-class combinations and contexts per s-class in the sampled
  Wiki-PSE \citep{yaghoobzadeh-etal-2019-probing}
  are listed in Table \ref{psestatisperclass}.
  Some annotated contexts in Wiki-PSE are also displayed in Table \ref{pseegs}.
  The Wiki-PSE developed by \citet{yaghoobzadeh-etal-2019-probing} is publicly available
  at \url{https://github.com/yyaghoobzadeh/WIKI-PSE}. 

  When finetuning BERT, we use the GLUE \citep{wang2018glue} splits of MRPC and SST2 from
  \url{https://gluebenchmark.com/}. Our POS dataset is from the linguistic data
  consortium (LDC). For NER \citep{nerconll2003}, we use the official shared task dataset:
  \url{https://www.clips.uantwerpen.be/conll2003/ner/}.

  \section{Finetuning Details}
  \seclabel{finetuningdetails}
  Hyperparameters in Table \ref{hypers} are used when we finetune BERT
  on POS, NER, SST2, and MRPC.
   For SST2 and MRPC, we use the embedding of 
   \texttt{[CLS]} as the representation of the sentence (pair).
   For POS and NER, we use the embedding of the last wordpiece of the word
    as \citet{liu-etal-2019-linguistic}.

    \begin{table}[]
      \small
      \begin{tabular}{c|cccc}
                          & POS  & SST2 & MRPC & NER\\ \hline
      batch size          & 150  & 200   & 350  & 32 \\
      learning rate       & 5e-5 & 5e-5  & 5e-5 & 5e-5\\
      max epoch           & 5    & 5     & 5    & 5\\
      max sequence length & 128  & 128   & 128  & 128 
      \end{tabular}
      \caption{Hyperparameters for finetuning.}
      \label{hypers}
      \end{table}
    
  A plain Adam \citep{kingma2014adam} optimizer is used and we did not use strategies like 
      learning rate warmup and layer-wise learning rate 
      \citep{howard-ruder-2018-universal} during finetuning
      to avoid potential side effects to ensure a clear
      comparison of different BERT layers.
  
  \section{Context Sizes in POS}
  \seclabel{ctxpos}
    
    We investigate how the findings from \secref{ctxlvl} in the main paper
    transfer to downstream tasks.  To this end we perform standard
    finetuning of BERT for different tasks, but we prune the attention
    matrix to a context size of length $k$. That is we apply a mask on
    the attention matrix such that each word can only attend to $k$
    left and $k$ right words. This has great benefits as it reduces
    the memory and computation requirements from $\mathcal{O}(n^2)$ to
    $\mathcal{O}(nk)$ where $n$ is the sequence length. We only
    consider part-of-speech tagging as for sentence pair
    classification tasks such as SST2 and MRPC this is not a sensible
    approach.
    
    Table \ref{windows} confirms that small context windows are
    sufficient to achieve full performance for POS-tagging. This
    indicates that the finding from the main paper (i.e., local
    context is sufficient for BERT to achieve a high degree of
    contextualization) is to some degree applicable to a downstream
    tasks, as well.  Note that the median sentence length in the Penn
    Treebank dataset is 25 words (the number of wordpieces even
    higher). Thus masking the context to the next 4 or 8 words
    does indeed reduce the available context words. In future work we
    plan to investigate this effect not only during finetuning but
    also during pretraining.

  \begin{table*}[t]
    \footnotesize
    \setlength{\tabcolsep}{4pt}
    \centering
    \begin{tabular}{@{}rrrrrr|rrrrrrrrrrrr@{}}
      \toprule
           & \multicolumn{5}{c|}{Standard Embeddings} & \multicolumn{12}{c}{AVG-BERT-$\ell$}                                                                                                               \\ \midrule
           & Rand                                     & BERTw                                & fastText & GloVe & PSE  & 0    & 1    & 2    & 3    & 4    & 5    & 6    & 7    & 8    & 9    & 10   & 11   \\ \midrule
      dev  & .269                                     & .653                                 & .625     & .681  & .790 & .746 & .759 & .764 & .775 & .786 & .791 & .794 & .805 & .811 & .812 & .813 & .809 \\
      test & .267                                     & .652                                 & .626     & .680  & .787 & .744 & .756 & .762 & .773 & .783 & .788 & .790 & .802 & .806 & .809 & .808 & .806 \\
      \bottomrule
    \end{tabular}
    \caption{S-class probing results
      for \textbf{uncontextualized} embeddings.
      Numbers are micro $F_1$ on Wiki-PSE.
      Our result
      (0.787 on PSE-test)
      is consistent with
      \citet{yaghoobzadeh-etal-2019-probing}.
      Additionally, for the top 6 layers \{6, 7, 8, 9, 10, 11\} of
      AVG-BERT, we repeat the experiments 5 times with random seed in
      \{1, 2, 3, 4, 5\}.  Mean and standard deviation on test per
      layer are: \{.791$\pm$.001, .801$\pm$.001,
      .807$\pm$.001, .808$\pm$.001, .808$\pm$.001,
      .805$\pm$.001\}.
    }
      \tablabel{uncontextualizedprob}
  \end{table*}
  
  \begin{table*}[t]
    \footnotesize
    \setlength{\tabcolsep}{5pt}
    \centering
    \begin{tabular}{@{}rrrrr|rrrrrrrrrrrr@{}}
      \toprule
           & \multicolumn{4}{c|}{Bag-of-word context} & \multicolumn{12}{c}{BERT Layer}                                                                                                           \\ \midrule
           &                                          & P-Rand                          & P-fastText & P-BERT & 0    & 1    & 2    & 3    & 4    & 5    & 6    & 7    & 8    & 9    & 10   & 11   \\ \midrule
      dev  &                                          & .637                            & .707       & .672   & .649 & .692 & .711 & .739 & .771 & .782 & .795 & .813 & .826 & .832 & .836 & .835 \\
      test &                                          & .630                            & .707       & .670   & .645 & .688 & .708 & .737 & .766 & .777 & .790 & .810 & .824 & .828 & .830 & .831 \\
      \bottomrule
    \end{tabular}
    \caption{S-class probing results for \textbf{contextualized} embedding models.
      Numbers are micro $F_1$ on Wiki-PSE.}
    \tablabel{contextualizedprob}
  \end{table*}

    \begin{table*}
      \centering
    	\begin{tabular}{c|c}
    		Context size & POS \\
    		\hline
    		0 & .886 \\
    		2 & .973 \\
    		4 & .975 \\
    		8 & .976 \\
    		16 & .977 \\
    		32 & .977 \\
    		All & .977 \\
    	\end{tabular}
    	\caption{POS accuracy on dev for different context sizes.}
    	\label{windows}
    \end{table*}

  \begin{table*}[t]
    \centering\small
    \begin{tabular}{r|rr|rr|rr|}
      \cline{2-7}
                                                              & \multicolumn{2}{c|}{train} & \multicolumn{2}{c|}{dev} & \multicolumn{2}{c|}{test} \\ \hline
      \multicolumn{1}{|r|}{semantic classes}                  & comb's     & contexts      & comb's     & contexts    & comb's     & contexts     \\ \hline
      \multicolumn{1}{|r|}{location}                          & 13,474     & 618,932       & 3,408      & 152,470     & 16,859     & 776,848      \\
      \multicolumn{1}{|r|}{person}                            & 15,423     & 617,270       & 3,744      & 151,005     & 19,212     & 765,655      \\
      \multicolumn{1}{|r|}{organization}                      & 9,556      & 332,063       & 2,496      & 88,682      & 11,915     & 411,716      \\
      \multicolumn{1}{|r|}{art}                               & 7,428      & 201,529       & 1,854      & 52,295      & 9,192      & 247,481      \\
      \multicolumn{1}{|r|}{event}                             & 3,515      & 87,735        & 900        & 21,566      & 4,404      & 108,963      \\
      \multicolumn{1}{|r|}{broadcast-program}                 & 2,287      & 67,261        & 530        & 15,062      & 2,828      & 84,343       \\
      \multicolumn{1}{|r|}{title}                             & 1,429      & 43,041        & 311        & 9,646       & 1,792      & 56,333       \\
      \multicolumn{1}{|r|}{product}                           & 3,121      & 49,076        & 766        & 13,438      & 3,808      & 61,585       \\
      \multicolumn{1}{|r|}{living-thing}                      & 1,302      & 35,595        & 320        & 9,035       & 1,702      & 46,040       \\
      \multicolumn{1}{|r|}{people-ethnicity}                  & 754        & 27,573        & 181        & 6,699       & 951        & 35,332       \\
      \multicolumn{1}{|r|}{language}                          & 671        & 14,842        & 145        & 3,147       & 824        & 20,308       \\
      \multicolumn{1}{|r|}{broadcast-network}                 & 325        & 12,392        & 80         & 3,036       & 362        & 13,006       \\
      \multicolumn{1}{|r|}{time}                              & 157        & 7,765         & 39         & 1,997       & 192        & 9,984        \\
      \multicolumn{1}{|r|}{religion-religion}                 & 192        & 6,461         & 45         & 1,760       & 265        & 9,719        \\
      \multicolumn{1}{|r|}{award}                             & 251        & 7,589         & 61         & 1,776       & 301        & 8,877        \\
      \multicolumn{1}{|r|}{internet-website}                  & 88         & 2,466         & 21         & 645         & 141        & 3,851        \\
      \multicolumn{1}{|r|}{god}                               & 246        & 7,306         & 52         & 1,998       & 340        & 11,810       \\
      \multicolumn{1}{|r|}{education-educational-degree}      & 97         & 3,282         & 24         & 901         & 142        & 4,833        \\
      \multicolumn{1}{|r|}{food}                              & 381        & 7,805         & 112        & 2,003       & 480        & 9,514        \\
      \multicolumn{1}{|r|}{computer-programming-language}     & 105        & 2,739         & 29         & 402         & 123        & 2,677        \\
      \multicolumn{1}{|r|}{metropolitan-transit-transit-line} & 285        & 5,603         & 76         & 1,259       & 382        & 6,948        \\
      \multicolumn{1}{|r|}{transit}                           & 135        & 3,781         & 26         & 628         & 186        & 4,305        \\
      \multicolumn{1}{|r|}{finance-currency}                  & 127        & 3,107         & 30         & 548         & 166        & 3,388        \\
      \multicolumn{1}{|r|}{disease}                           & 163        & 2,619         & 33         & 381         & 260        & 4,385        \\
      \multicolumn{1}{|r|}{chemistry}                         & 170        & 3,350         & 43         & 1,254       & 195        & 3,858        \\
      \multicolumn{1}{|r|}{body-part}                         & 135        & 1,901         & 31         & 415         & 156        & 2,591        \\
      \multicolumn{1}{|r|}{finance-stock-exchange}            & 27         & 617           & 3          & 5           & 51         & 795          \\
      \multicolumn{1}{|r|}{law}                               & 23         & 474           & 6          & 54          & 27         & 535          \\
      \multicolumn{1}{|r|}{medicine-medical-treatment}        & 77         & 886           & 7          & 124         & 106        & 1,803        \\
      \multicolumn{1}{|r|}{medicine-drug}                     & 50         & 1,023         & 7          & 54          & 72         & 1,157        \\
      \multicolumn{1}{|r|}{broadcast-tv-channel}              & 45         & 564           & 14         & 210         & 74         & 1,264        \\
      \multicolumn{1}{|r|}{medicine-symptom}                  & 55         & 752           & 15         & 97          & 72         & 1,172        \\
      \multicolumn{1}{|r|}{biology}                           & 49         & 485           & 15         & 118         & 63         & 911          \\
      \multicolumn{1}{|r|}{visual-art-color}                  & 41         & 1,011         & 13         & 228         & 63         & 906          \\ \hline
      \multicolumn{1}{|r|}{total}                             & 62,184     & 2,178,895     & 15,437     & 542,938     & 77,706     & 2,722,893    \\ \hline
      \end{tabular}
      \caption{Number of word-s-class combinations and contexts for each of the 34 semantic classes in Wiki-PSE.}
      \label{psestatisperclass}
    \end{table*}
  
    \begin{table*}[t]
      \scriptsize\centering
      \begin{tabular}{|c|c|l|}
        \hline
        word                     & word-s-class combination                & \multicolumn{1}{c|}{contexts}                                                                                       \\ \hline
        \multirow{6}{*}{roberta} & \multirow{2}{*}{@roberta@-art}          & this recording is also available on cd paired with @roberta@-art .                                                  \\
                                 &                                         & ... to star as huckleberry haines in the jerome kern / dorothy fields musical @roberta@-art .                       \\ \cline{2-3} 
                                 & \multirow{2}{*}{@roberta@-location}     & there are also learning centers in eatonton , forsyth , gray , jeffersonville , and @roberta@-location .            \\
                                 &                                         & ... the concurrency curves to a nearly due north routing and enters @roberta@-location .                            \\ \cline{2-3} 
                                 & \multirow{2}{*}{@roberta@-person}       & ken williams : along with wife @roberta@-person , founded on-line systems after working at ibm                      \\
                                 &                                         & mystery house is an adventure game released in 7 by @roberta@-person and ken williams for the apple ii .            \\ \hline
        \multirow{4}{*}{larch}   & \multirow{2}{*}{@larch@-comp-prog-lang} & wing has been a leading member of the formal methods community , especially in the area of @larch@-comp-prog-lang . \\
                                 &                                         & a major contribution was his involvement with the @larch@-comp-prog-lang approach to formal specification with ...  \\ \cline{2-3} 
                                 & \multirow{2}{*}{@larch@-living-thing}   & the more recent plantings include @larch@-living-thing and pine .                                                   \\
                                 &                                         & these consist mainly of oak , alder , @larch@-living-thing and corsican pine .                                      \\ \hline
        \end{tabular}
        \caption{Example contexts of the annotated word ``roberta'' and ``larch''.}
        \label{pseegs}
      \end{table*}

%% file: bert-pse.bbl
\begin{thebibliography}{69}
\expandafter\ifx\csname natexlab\endcsname\relax\def\natexlab#1{#1}\fi

\bibitem[{Adi et~al.(2016)Adi, Kermany, Belinkov, Lavi, and
  Goldberg}]{adi2016fine}
Yossi Adi, Einat Kermany, Yonatan Belinkov, Ofer Lavi, and Yoav Goldberg. 2016.
\newblock Fine-grained analysis of sentence embeddings using auxiliary
  prediction tasks.
\newblock \emph{arXiv preprint arXiv:1608.04207}.

\bibitem[{Artetxe et~al.(2018)Artetxe, Labaka, Lopez-Gazpio, and
  Agirre}]{artetxe-etal-2018-uncovering}
Mikel Artetxe, Gorka Labaka, I{\~n}igo Lopez-Gazpio, and Eneko Agirre. 2018.
\newblock \href {https://doi.org/10.18653/v1/K18-1028} {Uncovering divergent
  linguistic information in word embeddings with lessons for intrinsic and
  extrinsic evaluation}.
\newblock In \emph{Proceedings of the 22nd Conference on Computational Natural
  Language Learning}, pages 282--291, Brussels, Belgium. Association for
  Computational Linguistics.

\bibitem[{Belinkov and Glass(2019)}]{belinkov-glass-2019-analysis}
Yonatan Belinkov and James Glass. 2019.
\newblock \href {https://doi.org/10.1162/tacl_a_00254} {Analysis methods in
  neural language processing: A survey}.
\newblock \emph{Transactions of the Association for Computational Linguistics},
  7:49--72.

\bibitem[{Belinkov et~al.(2017)Belinkov, M{\`a}rquez, Sajjad, Durrani, Dalvi,
  and Glass}]{belinkov-etal-2017-evaluating}
Yonatan Belinkov, Llu{\'\i}s M{\`a}rquez, Hassan Sajjad, Nadir Durrani, Fahim
  Dalvi, and James Glass. 2017.
\newblock \href {https://www.aclweb.org/anthology/I17-1001} {Evaluating layers
  of representation in neural machine translation on part-of-speech and
  semantic tagging tasks}.
\newblock In \emph{Proceedings of the Eighth International Joint Conference on
  Natural Language Processing (Volume 1: Long Papers)}, pages 1--10, Taipei,
  Taiwan. Asian Federation of Natural Language Processing.

\bibitem[{Bevi{\'a} et~al.(2007)Bevi{\'a}, Cueto, and
  Claramunt}]{izquierdo2007exploring}
Rub{\'e}n~Izquierdo Bevi{\'a}, Armando~Su{\'a}rez Cueto, and Germ{\'a}n~Rigau
  Claramunt. 2007.
\newblock Exploring the automatic selection of basic level concepts.

\bibitem[{Black(1988)}]{black1988experiment}
Ezra Black. 1988.
\newblock An experiment in computational discrimination of english word senses.
\newblock \emph{IBM Journal of research and development}, 32(2):185--194.

\bibitem[{Bojanowski et~al.(2017)Bojanowski, Grave, Joulin, and
  Mikolov}]{bojanowski-etal-2017-enriching}
Piotr Bojanowski, Edouard Grave, Armand Joulin, and Tomas Mikolov. 2017.
\newblock \href {https://doi.org/10.1162/tacl_a_00051} {Enriching word vectors
  with subword information}.
\newblock \emph{Transactions of the Association for Computational Linguistics},
  5:135--146.

\bibitem[{Brunner et~al.(2020)Brunner, Liu, Pascual, Richter, Ciaramita, and
  Wattenhofer}]{Brunner2020On}
Gino Brunner, Yang Liu, Damian Pascual, Oliver Richter, Massimiliano Ciaramita,
  and Roger Wattenhofer. 2020.
\newblock \href {https://openreview.net/forum?id=BJg1f6EFDB} {On
  identifiability in transformers}.
\newblock In \emph{International Conference on Learning Representations}.

\bibitem[{Ciaramita and Johnson(2003)}]{ciaramita2003supersense}
Massimiliano Ciaramita and Mark Johnson. 2003.
\newblock Supersense tagging of unknown nouns in wordnet.
\newblock In \emph{Proceedings of the 2003 conference on Empirical methods in
  natural language processing}, pages 168--175. Association for Computational
  Linguistics.

\bibitem[{Clark et~al.(2019)Clark, Khandelwal, Levy, and
  Manning}]{clark-etal-2019-bert}
Kevin Clark, Urvashi Khandelwal, Omer Levy, and Christopher~D. Manning. 2019.
\newblock \href {https://doi.org/10.18653/v1/W19-4828} {What does {BERT} look
  at? an analysis of {BERT}{'}s attention}.
\newblock In \emph{Proceedings of the 2019 ACL Workshop BlackboxNLP: Analyzing
  and Interpreting Neural Networks for NLP}, pages 276--286, Florence, Italy.
  Association for Computational Linguistics.

\bibitem[{Collins(2002)}]{collins-2002-discriminative}
Michael Collins. 2002.
\newblock \href {https://doi.org/10.3115/1118693.1118694} {Discriminative
  training methods for hidden {M}arkov models: Theory and experiments with
  perceptron algorithms}.
\newblock In \emph{Proceedings of the 2002 Conference on Empirical Methods in
  Natural Language Processing ({EMNLP} 2002)}, pages 1--8. Association for
  Computational Linguistics.

\bibitem[{Conneau et~al.(2018)Conneau, Kruszewski, Lample, Barrault, and
  Baroni}]{conneau-etal-2018-cram}
Alexis Conneau, German Kruszewski, Guillaume Lample, Lo{\"\i}c Barrault, and
  Marco Baroni. 2018.
\newblock \href {https://doi.org/10.18653/v1/P18-1198} {What you can cram into
  a single {\$}{\&}!{\#}* vector: Probing sentence embeddings for linguistic
  properties}.
\newblock In \emph{Proceedings of the 56th Annual Meeting of the Association
  for Computational Linguistics (Volume 1: Long Papers)}, pages 2126--2136,
  Melbourne, Australia. Association for Computational Linguistics.

\bibitem[{Dai and Le(2015)}]{dai2015semisupervised}
Andrew~M. Dai and Quoc~V. Le. 2015.
\newblock \href {http://arxiv.org/abs/1511.01432} {Semi-supervised sequence
  learning}.

\bibitem[{Devlin et~al.(2019)Devlin, Chang, Lee, and
  Toutanova}]{devlin-etal-2019-bert}
Jacob Devlin, Ming-Wei Chang, Kenton Lee, and Kristina Toutanova. 2019.
\newblock \href {https://doi.org/10.18653/v1/N19-1423} {{BERT}: Pre-training of
  deep bidirectional transformers for language understanding}.
\newblock In \emph{Proceedings of the 2019 Conference of the North {A}merican
  Chapter of the Association for Computational Linguistics: Human Language
  Technologies, Volume 1 (Long and Short Papers)}, pages 4171--4186,
  Minneapolis, Minnesota. Association for Computational Linguistics.

\bibitem[{Dolan and Brockett(2005)}]{dolan2005automatically}
William~B Dolan and Chris Brockett. 2005.
\newblock Automatically constructing a corpus of sentential paraphrases.
\newblock In \emph{Proceedings of the Third International Workshop on
  Paraphrasing (IWP2005)}.

\bibitem[{Edmonds and Cotton(2001)}]{edmonds-cotton-2001-senseval}
Philip Edmonds and Scott Cotton. 2001.
\newblock \href {https://www.aclweb.org/anthology/S01-1001} {{SENSEVAL}-2:
  Overview}.
\newblock In \emph{Proceedings of {SENSEVAL}-2 Second International Workshop on
  Evaluating Word Sense Disambiguation Systems}, pages 1--5, Toulouse, France.
  Association for Computational Linguistics.

\bibitem[{Ethayarajh(2019)}]{ethayarajh2019contextual}
Kawin Ethayarajh. 2019.
\newblock \href {https://www.aclweb.org/anthology/D19-1006} {How contextual are
  contextualized word representations? comparing the geometry of {BERT},
  {ELM}o, and {GPT}-2 embeddings}.
\newblock In \emph{Proceedings of the 2019 Conference on Empirical Methods in
  Natural Language Processing and the 9th International Joint Conference on
  Natural Language Processing (EMNLP-IJCNLP)}, pages 55--65, Hong Kong, China.
  Association for Computational Linguistics.

\bibitem[{Fellbaum(1998)}]{wordnetcite}
Christiane Fellbaum. 1998.
\newblock \emph{WordNet: An Electronic Lexical Database}.
\newblock Bradford Books.

\bibitem[{Gale et~al.(1992)Gale, Church, and Yarowsky}]{gale1992one}
William~A Gale, Kenneth~W Church, and David Yarowsky. 1992.
\newblock One sense per discourse.
\newblock In \emph{Proceedings of the workshop on Speech and Natural Language},
  pages 233--237. Association for Computational Linguistics.

\bibitem[{Goldberg(2019)}]{goldberg2019assessing}
Yoav Goldberg. 2019.
\newblock Assessing bert's syntactic abilities.
\newblock \emph{arXiv preprint arXiv:1901.05287}.

\bibitem[{Hao et~al.(2019)Hao, Dong, Wei, and Xu}]{hao2019visualizing}
Yaru Hao, Li~Dong, Furu Wei, and Ke~Xu. 2019.
\newblock \href {https://www.aclweb.org/anthology/D19-1424} {Visualizing and
  understanding the effectiveness of {BERT}}.
\newblock In \emph{Proceedings of the 2019 Conference on Empirical Methods in
  Natural Language Processing and the 9th International Joint Conference on
  Natural Language Processing (EMNLP-IJCNLP)}, pages 4134--4143, Hong Kong,
  China. Association for Computational Linguistics.

\bibitem[{He and Choi(2020)}]{he2019establishing}
Han He and Jinho~D. Choi. 2020.
\newblock \href {https://www.flairs-33.info} {{Establishing Strong Baselines
  for the New Decade: Sequence Tagging, Syntactic and Semantic Parsing with
  BERT}}.
\newblock In \emph{Proceedings of the 33rd International Florida Artificial
  Intelligence Research Society Conference}, FLAIRS'20.
\newblock Best Paper Candidate.

\bibitem[{Hewitt and Liang(2019)}]{hewitt-liang-2019-designing}
John Hewitt and Percy Liang. 2019.
\newblock \href {https://doi.org/10.18653/v1/D19-1275} {Designing and
  interpreting probes with control tasks}.
\newblock In \emph{Proceedings of the 2019 Conference on Empirical Methods in
  Natural Language Processing and the 9th International Joint Conference on
  Natural Language Processing (EMNLP-IJCNLP)}, pages 2733--2743, Hong Kong,
  China. Association for Computational Linguistics.

\bibitem[{Hochreiter and Schmidhuber(1997)}]{hochreiter1997long}
Sepp Hochreiter and J{\"u}rgen Schmidhuber. 1997.
\newblock Long short-term memory.
\newblock \emph{Neural computation}, 9(8):1735--1780.

\bibitem[{Howard and Ruder(2018)}]{howard-ruder-2018-universal}
Jeremy Howard and Sebastian Ruder. 2018.
\newblock \href {https://doi.org/10.18653/v1/P18-1031} {Universal language
  model fine-tuning for text classification}.
\newblock In \emph{Proceedings of the 56th Annual Meeting of the Association
  for Computational Linguistics (Volume 1: Long Papers)}, pages 328--339,
  Melbourne, Australia. Association for Computational Linguistics.

\bibitem[{Izquierdo et~al.(2009)Izquierdo, Su{\'a}rez, and
  Rigau}]{izquierdo2009empirical}
Rub{\'e}n Izquierdo, Armando Su{\'a}rez, and German Rigau. 2009.
\newblock An empirical study on class-based word sense disambiguation.
\newblock In \emph{Proceedings of the 12th Conference of the European Chapter
  of the Association for Computational Linguistics}, pages 389--397.
  Association for Computational Linguistics.

\bibitem[{Jurafsky and Martin(2009)}]{Jurafsky:2009:SLP:1214993}
Daniel Jurafsky and James~H. Martin. 2009.
\newblock \emph{Speech and Language Processing (2Nd Edition)}.
\newblock Prentice-Hall, Inc., Upper Saddle River, NJ, USA.

\bibitem[{Kingma and Ba(2014)}]{kingma2014adam}
Diederik~P Kingma and Jimmy Ba. 2014.
\newblock Adam: A method for stochastic optimization.
\newblock \emph{arXiv preprint arXiv:1412.6980}.

\bibitem[{Kohomban and Lee(2005)}]{kohomban-lee-2005-learning}
Upali~Sathyajith Kohomban and Wee~Sun Lee. 2005.
\newblock \href {https://doi.org/10.3115/1219840.1219845} {Learning semantic
  classes for word sense disambiguation}.
\newblock In \emph{Proceedings of the 43rd Annual Meeting of the Association
  for Computational Linguistics ({ACL}{'}05)}, pages 34--41, Ann Arbor,
  Michigan. Association for Computational Linguistics.

\bibitem[{Kovaleva et~al.(2019)Kovaleva, Romanov, Rogers, and
  Rumshisky}]{kovaleva2019revealing}
Olga Kovaleva, Alexey Romanov, Anna Rogers, and Anna Rumshisky. 2019.
\newblock \href {https://www.aclweb.org/anthology/D19-1445} {Revealing the dark
  secrets of {BERT}}.
\newblock In \emph{Proceedings of the 2019 Conference on Empirical Methods in
  Natural Language Processing and the 9th International Joint Conference on
  Natural Language Processing (EMNLP-IJCNLP)}, pages 4356--4365, Hong Kong,
  China. Association for Computational Linguistics.

\bibitem[{Lee et~al.(2019)Lee, Tang, and Lin}]{lee2019elsa}
Jaejun Lee, Raphael Tang, and Jimmy Lin. 2019.
\newblock \href {http://arxiv.org/abs/1911.03090} {What would elsa do? freezing
  layers during transformer fine-tuning}.

\bibitem[{Levine et~al.(2019)Levine, Lenz, Dagan, Padnos, Sharir,
  Shalev-Shwartz, Shashua, and Shoham}]{levine2019sensebert}
Yoav Levine, Barak Lenz, Or~Dagan, Dan Padnos, Or~Sharir, Shai Shalev-Shwartz,
  Amnon Shashua, and Yoav Shoham. 2019.
\newblock Sensebert: Driving some sense into bert.
\newblock \emph{arXiv preprint arXiv:1908.05646}.

\bibitem[{Lin et~al.(2019)Lin, Tan, and Frank}]{lin2019open}
Yongjie Lin, Yi~Chern Tan, and Robert Frank. 2019.
\newblock Open sesame: Getting inside bert's linguistic knowledge.
\newblock \emph{arXiv preprint arXiv:1906.01698}.

\bibitem[{Linzen et~al.(2016)Linzen, Dupoux, and
  Goldberg}]{linzen-etal-2016-assessing}
Tal Linzen, Emmanuel Dupoux, and Yoav Goldberg. 2016.
\newblock \href {https://doi.org/10.1162/tacl_a_00115} {Assessing the ability
  of {LSTM}s to learn syntax-sensitive dependencies}.
\newblock \emph{Transactions of the Association for Computational Linguistics},
  4:521--535.

\bibitem[{Liu et~al.(2019)Liu, Gardner, Belinkov, Peters, and
  Smith}]{liu-etal-2019-linguistic}
Nelson~F. Liu, Matt Gardner, Yonatan Belinkov, Matthew~E. Peters, and Noah~A.
  Smith. 2019.
\newblock \href {https://doi.org/10.18653/v1/N19-1112} {Linguistic knowledge
  and transferability of contextual representations}.
\newblock In \emph{Proceedings of the 2019 Conference of the North {A}merican
  Chapter of the Association for Computational Linguistics: Human Language
  Technologies, Volume 1 (Long and Short Papers)}, pages 1073--1094,
  Minneapolis, Minnesota. Association for Computational Linguistics.

\bibitem[{Magnini and Cavagli{\`a}(2000)}]{magnini-cavaglia-2000-integrating}
Bernardo Magnini and Gabriela Cavagli{\`a}. 2000.
\newblock \href {http://www.lrec-conf.org/proceedings/lrec2000/pdf/219.pdf}
  {Integrating subject field codes into {W}ord{N}et}.
\newblock In \emph{Proceedings of the Second International Conference on
  Language Resources and Evaluation ({LREC}{'}00)}, Athens, Greece. European
  Language Resources Association (ELRA).

\bibitem[{Marcus et~al.(1993)Marcus, Santorini, and
  Marcinkiewicz}]{marcus1993building}
Mitchell~P. Marcus, Beatrice Santorini, and Mary~Ann Marcinkiewicz. 1993.
\newblock \href {https://www.aclweb.org/anthology/J93-2004} {Building a large
  annotated corpus of {E}nglish: The {P}enn {T}reebank}.
\newblock \emph{Computational Linguistics}, 19(2):313--330.

\bibitem[{McCann et~al.(2017)McCann, Bradbury, Xiong, and
  Socher}]{mccann2017learned}
Bryan McCann, James Bradbury, Caiming Xiong, and Richard Socher. 2017.
\newblock Learned in translation: Contextualized word vectors.
\newblock In \emph{Advances in Neural Information Processing Systems}, pages
  6294--6305.

\bibitem[{Merchant et~al.(2020)Merchant, Rahimtoroghi, Pavlick, and
  Tenney}]{merchant2020happens}
Amil Merchant, Elahe Rahimtoroghi, Ellie Pavlick, and Ian Tenney. 2020.
\newblock \href {http://arxiv.org/abs/2004.14448} {What happens to bert
  embeddings during fine-tuning?}

\bibitem[{Mikolov et~al.(2013)Mikolov, Sutskever, Chen, Corrado, and
  Dean}]{mikolov2013distributed}
Tomas Mikolov, Ilya Sutskever, Kai Chen, Greg~S Corrado, and Jeff Dean. 2013.
\newblock Distributed representations of words and phrases and their
  compositionality.
\newblock In \emph{Advances in neural information processing systems}, pages
  3111--3119.

\bibitem[{Paszke et~al.(2019)Paszke, Gross, Massa, Lerer, Bradbury, Chanan,
  Killeen, Lin, Gimelshein, Antiga, Desmaison, Kopf, Yang, DeVito, Raison,
  Tejani, Chilamkurthy, Steiner, Fang, Bai, and Chintala}]{paszke2017automatic}
Adam Paszke, Sam Gross, Francisco Massa, Adam Lerer, James Bradbury, Gregory
  Chanan, Trevor Killeen, Zeming Lin, Natalia Gimelshein, Luca Antiga, Alban
  Desmaison, Andreas Kopf, Edward Yang, Zachary DeVito, Martin Raison, Alykhan
  Tejani, Sasank Chilamkurthy, Benoit Steiner, Lu~Fang, Junjie Bai, and Soumith
  Chintala. 2019.
\newblock \href
  {http://papers.nips.cc/paper/9015-pytorch-an-imperative-style-high-performance-deep-learning-library.pdf}
  {Pytorch: An imperative style, high-performance deep learning library}.
\newblock In \emph{Advances in Neural Information Processing Systems 32}, pages
  8024--8035. Curran Associates, Inc.

\bibitem[{Pennington et~al.(2014)Pennington, Socher, and
  Manning}]{Pennington2014}
Jeffrey Pennington, Richard Socher, and Christopher Manning. 2014.
\newblock \href {https://doi.org/10.3115/v1/D14-1162} {{Glove: Global Vectors
  for Word Representation}}.
\newblock \emph{Proceedings of the 2014 Conference on Empirical Methods in
  Natural Language Processing (EMNLP)}, pages 1532--1543.

\bibitem[{Peters et~al.(2018{\natexlab{a}})Peters, Neumann, Iyyer, Gardner,
  Clark, Lee, and Zettlemoyer}]{peters-etal-2018-deep}
Matthew Peters, Mark Neumann, Mohit Iyyer, Matt Gardner, Christopher Clark,
  Kenton Lee, and Luke Zettlemoyer. 2018{\natexlab{a}}.
\newblock \href {https://doi.org/10.18653/v1/N18-1202} {Deep contextualized
  word representations}.
\newblock In \emph{Proceedings of the 2018 Conference of the North {A}merican
  Chapter of the Association for Computational Linguistics: Human Language
  Technologies, Volume 1 (Long Papers)}, pages 2227--2237, New Orleans,
  Louisiana. Association for Computational Linguistics.

\bibitem[{Peters et~al.(2018{\natexlab{b}})Peters, Neumann, Zettlemoyer, and
  Yih}]{peters-etal-2018-dissecting}
Matthew Peters, Mark Neumann, Luke Zettlemoyer, and Wen-tau Yih.
  2018{\natexlab{b}}.
\newblock \href {https://doi.org/10.18653/v1/D18-1179} {Dissecting contextual
  word embeddings: Architecture and representation}.
\newblock In \emph{Proceedings of the 2018 Conference on Empirical Methods in
  Natural Language Processing}, pages 1499--1509, Brussels, Belgium.
  Association for Computational Linguistics.

\bibitem[{Peters et~al.(2019)Peters, Ruder, and Smith}]{peters-etal-2019-tune}
Matthew~E. Peters, Sebastian Ruder, and Noah~A. Smith. 2019.
\newblock \href {https://doi.org/10.18653/v1/W19-4302} {To tune or not to tune?
  adapting pretrained representations to diverse tasks}.
\newblock In \emph{Proceedings of the 4th Workshop on Representation Learning
  for NLP (RepL4NLP-2019)}, pages 7--14, Florence, Italy. Association for
  Computational Linguistics.

\bibitem[{Pinter et~al.(2017)Pinter, Guthrie, and
  Eisenstein}]{pinter-etal-2017-mimicking}
Yuval Pinter, Robert Guthrie, and Jacob Eisenstein. 2017.
\newblock \href {https://doi.org/10.18653/v1/D17-1010} {Mimicking word
  embeddings using subword {RNN}s}.
\newblock In \emph{Proceedings of the 2017 Conference on Empirical Methods in
  Natural Language Processing}, pages 102--112, Copenhagen, Denmark.
  Association for Computational Linguistics.

\bibitem[{Radford et~al.(2019)Radford, Wu, Child, Luan, Amodei, and
  Sutskever}]{radford2019language}
Alec Radford, Jeffrey Wu, Rewon Child, David Luan, Dario Amodei, and Ilya
  Sutskever. 2019.
\newblock Language models are unsupervised multitask learners.
\newblock \emph{OpenAI Blog}, 1(8).

\bibitem[{Raganato et~al.(2017)Raganato, Camacho-Collados, and
  Navigli}]{raganato-etal-2017-word}
Alessandro Raganato, Jose Camacho-Collados, and Roberto Navigli. 2017.
\newblock \href {https://www.aclweb.org/anthology/E17-1010} {Word sense
  disambiguation: A unified evaluation framework and empirical comparison}.
\newblock In \emph{Proceedings of the 15th Conference of the {E}uropean Chapter
  of the Association for Computational Linguistics: Volume 1, Long Papers},
  pages 99--110, Valencia, Spain. Association for Computational Linguistics.

\bibitem[{Resnik(1993)}]{resnik-1993-semantic}
Philip Resnik. 1993.
\newblock \href {https://www.aclweb.org/anthology/H93-1054} {Semantic classes
  and syntactic ambiguity}.
\newblock In \emph{{HUMAN} {LANGUAGE} {TECHNOLOGY}: Proceedings of a Workshop
  Held at Plainsboro, New Jersey, March 21-24, 1993}.

\bibitem[{Schick and Sch{\"{u}}tze(2020)}]{schick2019rare}
Timo Schick and Hinrich Sch{\"{u}}tze. 2020.
\newblock \href {https://aaai.org/ojs/index.php/AAAI/article/view/6403} {Rare
  words: {A} major problem for contextualized embeddings and how to fix it by
  attentive mimicking}.
\newblock In \emph{The Thirty-Fourth {AAAI} Conference on Artificial
  Intelligence, {AAAI} 2020, New York, NY, USA, February 7-12, 2020}, pages
  8766--8774. {AAAI} Press.

\bibitem[{Schuster et~al.(2019)Schuster, Ram, Barzilay, and
  Globerson}]{schuster-etal-2019-cross}
Tal Schuster, Ori Ram, Regina Barzilay, and Amir Globerson. 2019.
\newblock \href {https://doi.org/10.18653/v1/N19-1162} {Cross-lingual alignment
  of contextual word embeddings, with applications to zero-shot dependency
  parsing}.
\newblock In \emph{Proceedings of the 2019 Conference of the North {A}merican
  Chapter of the Association for Computational Linguistics: Human Language
  Technologies, Volume 1 (Long and Short Papers)}, pages 1599--1613,
  Minneapolis, Minnesota. Association for Computational Linguistics.

\bibitem[{Schwartz et~al.(2019)Schwartz, Dodge, Smith, and
  Etzioni}]{schwartz2019green}
Roy Schwartz, Jesse Dodge, Noah~A. Smith, and Oren Etzioni. 2019.
\newblock \href {http://arxiv.org/abs/1907.10597} {Green ai}.

\bibitem[{Shi et~al.(2016)Shi, Padhi, and Knight}]{shi-etal-2016-string}
Xing Shi, Inkit Padhi, and Kevin Knight. 2016.
\newblock \href {https://doi.org/10.18653/v1/D16-1159} {Does string-based
  neural {MT} learn source syntax?}
\newblock In \emph{Proceedings of the 2016 Conference on Empirical Methods in
  Natural Language Processing}, pages 1526--1534, Austin, Texas. Association
  for Computational Linguistics.

\bibitem[{Snyder and Palmer(2004)}]{snyder-palmer-2004-english}
Benjamin Snyder and Martha Palmer. 2004.
\newblock \href {https://www.aclweb.org/anthology/W04-0811} {The {E}nglish
  all-words task}.
\newblock In \emph{Proceedings of {SENSEVAL}-3, the Third International
  Workshop on the Evaluation of Systems for the Semantic Analysis of Text},
  pages 41--43, Barcelona, Spain. Association for Computational Linguistics.

\bibitem[{Socher et~al.(2013)Socher, Perelygin, Wu, Chuang, Manning, Ng, and
  Potts}]{socher2013recursive}
Richard Socher, Alex Perelygin, Jean Wu, Jason Chuang, Christopher~D Manning,
  Andrew Ng, and Christopher Potts. 2013.
\newblock Recursive deep models for semantic compositionality over a sentiment
  treebank.
\newblock In \emph{Proceedings of the 2013 conference on empirical methods in
  natural language processing}, pages 1631--1642.

\bibitem[{Strubell et~al.(2019)Strubell, Ganesh, and
  McCallum}]{strubell-etal-2019-energy}
Emma Strubell, Ananya Ganesh, and Andrew McCallum. 2019.
\newblock \href {https://doi.org/10.18653/v1/P19-1355} {Energy and policy
  considerations for deep learning in {NLP}}.
\newblock In \emph{Proceedings of the 57th Annual Meeting of the Association
  for Computational Linguistics}, pages 3645--3650, Florence, Italy.
  Association for Computational Linguistics.

\bibitem[{Tenney et~al.(2019{\natexlab{a}})Tenney, Das, and
  Pavlick}]{tenney-etal-2019-bert}
Ian Tenney, Dipanjan Das, and Ellie Pavlick. 2019{\natexlab{a}}.
\newblock \href {https://www.aclweb.org/anthology/P19-1452} {{BERT} rediscovers
  the classical {NLP} pipeline}.
\newblock In \emph{Proceedings of the 57th Annual Meeting of the Association
  for Computational Linguistics}, pages 4593--4601, Florence, Italy.
  Association for Computational Linguistics.

\bibitem[{Tenney et~al.(2019{\natexlab{b}})Tenney, Xia, Chen, Wang, Poliak,
  McCoy, Kim, Van~Durme, Bowman, Das et~al.}]{tenney2019you}
Ian Tenney, Patrick Xia, Berlin Chen, Alex Wang, Adam Poliak, R~Thomas McCoy,
  Najoung Kim, Benjamin Van~Durme, Samuel~R Bowman, Dipanjan Das, et~al.
  2019{\natexlab{b}}.
\newblock What do you learn from context? probing for sentence structure in
  contextualized word representations.
\newblock \emph{arXiv preprint arXiv:1905.06316}.

\bibitem[{Tjong Kim~Sang and De~Meulder(2003)}]{nerconll2003}
Erik~F. Tjong Kim~Sang and Fien De~Meulder. 2003.
\newblock \href {https://www.aclweb.org/anthology/W03-0419} {Introduction to
  the {C}o{NLL}-2003 shared task: Language-independent named entity
  recognition}.
\newblock In \emph{Proceedings of the Seventh Conference on Natural Language
  Learning at {HLT}-{NAACL} 2003}, pages 142--147.

\bibitem[{Vaswani et~al.(2017)Vaswani, Shazeer, Parmar, Uszkoreit, Jones,
  Gomez, Kaiser, and Polosukhin}]{vaswani2017attention}
Ashish Vaswani, Noam Shazeer, Niki Parmar, Jakob Uszkoreit, Llion Jones,
  Aidan~N Gomez, {\L}ukasz Kaiser, and Illia Polosukhin. 2017.
\newblock Attention is all you need.
\newblock In \emph{Advances in neural information processing systems}, pages
  5998--6008.

\bibitem[{Voita et~al.(2019)Voita, Sennrich, and
  Titov}]{voita-etal-2019-bottom}
Elena Voita, Rico Sennrich, and Ivan Titov. 2019.
\newblock \href {https://doi.org/10.18653/v1/D19-1448} {The bottom-up evolution
  of representations in the transformer: A study with machine translation and
  language modeling objectives}.
\newblock In \emph{Proceedings of the 2019 Conference on Empirical Methods in
  Natural Language Processing and the 9th International Joint Conference on
  Natural Language Processing (EMNLP-IJCNLP)}, pages 4387--4397, Hong Kong,
  China. Association for Computational Linguistics.

\bibitem[{Wang et~al.(2019)Wang, Pruksachatkun, Nangia, Singh, Michael, Hill,
  Levy, and Bowman}]{wang2019superglue}
Alex Wang, Yada Pruksachatkun, Nikita Nangia, Amanpreet Singh, Julian Michael,
  Felix Hill, Omer Levy, and Samuel~R Bowman. 2019.
\newblock Superglue: A stickier benchmark for general-purpose language
  understanding systems.
\newblock \emph{arXiv preprint arXiv:1905.00537}.

\bibitem[{Wang et~al.(2018)Wang, Singh, Michael, Hill, Levy, and
  Bowman}]{wang2018glue}
Alex Wang, Amanpreet Singh, Julian Michael, Felix Hill, Omer Levy, and Samuel~R
  Bowman. 2018.
\newblock Glue: A multi-task benchmark and analysis platform for natural
  language understanding.
\newblock \emph{arXiv preprint arXiv:1804.07461}.

\bibitem[{Wieting and Kiela(2019)}]{wieting2019no}
John Wieting and Douwe Kiela. 2019.
\newblock No training required: Exploring random encoders for sentence
  classification.
\newblock \emph{arXiv preprint arXiv:1901.10444}.

\bibitem[{Wolf et~al.(2019)Wolf, Debut, Sanh, Chaumond, Delangue, Moi, Cistac,
  Rault, Louf, Funtowicz, and Brew}]{wolf2019transformers}
Thomas Wolf, Lysandre Debut, Victor Sanh, Julien Chaumond, Clement Delangue,
  Anthony Moi, Pierric Cistac, Tim Rault, Rémi Louf, Morgan Funtowicz, and
  Jamie Brew. 2019.
\newblock \href {http://arxiv.org/abs/1910.03771} {Transformers:
  State-of-the-art natural language processing}.

\bibitem[{Wu et~al.(2016)Wu, Schuster, Chen, Le, Norouzi, Macherey, Krikun,
  Cao, Gao, Macherey et~al.}]{wu2016google}
Yonghui Wu, Mike Schuster, Zhifeng Chen, Quoc~V Le, Mohammad Norouzi, Wolfgang
  Macherey, Maxim Krikun, Yuan Cao, Qin Gao, Klaus Macherey, et~al. 2016.
\newblock Google's neural machine translation system: Bridging the gap between
  human and machine translation.
\newblock \emph{arXiv preprint arXiv:1609.08144}.

\bibitem[{Yaghoobzadeh et~al.(2019)Yaghoobzadeh, Kann, Hazen, Agirre, and
  Sch{\"u}tze}]{yaghoobzadeh-etal-2019-probing}
Yadollah Yaghoobzadeh, Katharina Kann, T.~J. Hazen, Eneko Agirre, and Hinrich
  Sch{\"u}tze. 2019.
\newblock \href {https://www.aclweb.org/anthology/P19-1574} {Probing for
  semantic classes: Diagnosing the meaning content of word embeddings}.
\newblock In \emph{Proceedings of the 57th Annual Meeting of the Association
  for Computational Linguistics}, pages 5740--5753, Florence, Italy.
  Association for Computational Linguistics.

\bibitem[{Yang et~al.(2019)Yang, Dai, Yang, Carbonell, Salakhutdinov, and
  Le}]{Yang2019XLNetGA}
Zhilin Yang, Zihang Dai, Yiming Yang, Jaime~G. Carbonell, Ruslan Salakhutdinov,
  and Quoc~V. Le. 2019.
\newblock Xlnet: Generalized autoregressive pretraining for language
  understanding.
\newblock \emph{ArXiv}, abs/1906.08237.

\bibitem[{Yarowsky(1992)}]{yarowsky-1992-word}
David Yarowsky. 1992.
\newblock \href {https://www.aclweb.org/anthology/C92-2070} {Word-sense
  disambiguation using statistical models of {R}oget{'}s categories trained on
  large corpora}.
\newblock In \emph{{COLING} 1992 Volume 2: The 15th International Conference on
  Computational Linguistics}.

\end{thebibliography}
